\documentclass[12pt,journal,a4paper,onecolumn]{IEEEtran}

\usepackage{setspace} 
\usepackage{dsfont}
\usepackage{amsmath}
\usepackage{amssymb}
\usepackage{amsfonts}
\usepackage{graphicx}
\usepackage{amsmath}
\usepackage[all,poly]{xy}
\usepackage{multirow}
\usepackage{color}
\usepackage[noadjust]{cite}
\usepackage{subfigure}
\usepackage{mathrsfs}
\usepackage[linesnumbered,lined,ruled]{algorithm2e}
\usepackage{algorithmicx}
\usepackage{cancel}
\usepackage{url}
\usepackage{lipsum}
\usepackage{stfloats}
\usepackage{epsfig}

\def\bfa{{\mathbf{a}}}

\def\bfn{{\mathbf{n}}}

\def\bfp{{\mathbf{p}}}

\def\bfy{{\mathbf{y}}}

\def\bfA{{\mathbf{A}}}
\def\bfB{{\mathbf{B}}}
\def\bfC{{\mathbf{C}}}
\def\bfD{{\mathbf{D}}}

\def\bfG{{\mathbf{G}}}
\def\bfH{{\mathbf{H}}}

\def\bfM{{\mathbf{M}}}
\def\bfN{{\mathbf{N}}}

\def\bfR{{\mathbf{R}}}
\def\bfS{{\mathbf{S}}}
\def\bfT{{\mathbf{T}}}
\def\bfU{{\mathbf{U}}}
\def\bfV{{\mathbf{V}}}

\def\bfX{{\mathbf{X}}}
\def\bfY{{\mathbf{Y}}}

\def\calP{{\mathcal{P}}}

\def\calN{{\mathcal{N}}}

\def\calP{{\mathcal{P}}}

\def\bsu{{\boldsymbol{u}}}

\def\bsx{{\boldsymbol{x}}}
\def\bsy{{\boldsymbol{y}}}

\def\wtm{\widetilde{m}}

\newcommand{\MATima}{\bfX}

\newcommand{\noisevar}[1]{{s^2_{#1}}}

\newcommand{\nbbandima}{m_{\lambda}}

\newcommand{\argmin}{\mathrm{arg}\min}

\newcommand{\Vzeros}[1]{\boldsymbol{0}_{#1}}
\newcommand{\Id}[1]{\textbf{I}_{#1}}

\newcounter{algo}
\renewcommand{\thealgo}{\arabic{algo}}

\graphicspath{{figures/}}

\newcommand{\bs}{\boldsymbol}

\newcommand{\NoiCovMat}{\bs{\Lambda}}

\title{Hyperspectral and Multispectral Image Fusion based on a Sparse Representation }

\author{\IEEEauthorblockN{Qi Wei, \IEEEmembership{Student Member,~IEEE},
Jos\'e Bioucas-Dias, \IEEEmembership{Senior Member,~IEEE},\\
Nicolas Dobigeon, \IEEEmembership{Senior Member,~IEEE}, and Jean-Yves
Tourneret, \IEEEmembership{Senior Member,~IEEE}}
\thanks{This work has been supported in part by the Hypanema ANR Project
n$^\circ$ANR-12-BS03-003, by ANR-11-LABX-0040-CIMI within the program
ANR-11-IDEX-0002-02 within the thematic trimester on image processing,
by the Portuguese Science and Technology Foundation under
Projects PEst-OE/EEI/LA0008/2013 and PTDC/EEI-PRO/1470/2012,
and Chinese Scholarship Council. Part of this work was presented 
in Proceedings of the 22nd European Signal Processing 
Conference (EUSIPCO), 2014 \cite{Wei2014Eusipco}.}

\thanks{Qi Wei, Nicolas Dobigeon and Jean-Yves
Tourneret are with IRIT/INP-ENSEEIHT, University of 
Toulouse, Toulouse, France (e-mail: \{qi.wei, nicolas.dobigeon, 
jean-yves.tourneret \}@enseeiht.fr) and Jos\'e Bioucas-Dias 
is with Instituto de Telecomunica\c{c}\~oes and Instituto Superior T\'ecnico, Universidade de Lisboa, Portugal.}}

\begin{document}
\maketitle
\hyphenation{hie-rar-chi-cal as-tro-no-mi-cal}

\begin{abstract}
This paper presents a variational based approach to fusing hyperspectral and multispectral images. 
The fusion process is formulated as an inverse problem whose solution is the target image
assumed to live in a much lower dimensional subspace. A sparse regularization term is carefully designed,
relying on a decomposition of the scene on a set of dictionaries. The dictionary atoms and the corresponding supports of active coding coefficients are learned from the observed images.
Then, conditionally on these dictionaries and supports, the fusion problem is solved via alternating optimization
with respect to the target image (using the alternating direction method of multipliers)
and the coding coefficients. Simulation results demonstrate the efficiency of the proposed
algorithm when compared with the state-of-the-art fusion methods.
\end{abstract}


\section{Introduction}
\label{sec:intro}

Fusion of multi-sensor images has been explored during recent years and is still a very active research area
\cite{Amro2011survey}. A popular fusion problem  in remote sensing consists of merging a high spatial resolution panchromatic (PAN) image and a low spatial resolution multispectral (MS) image. Many solutions have been proposed in the literature to solve this problem, known as \emph{pansharpening} \cite{Amro2011survey,Gonzalez2004fusion,Li2011,Liu2012}.

More recently, hyperspectral (HS) imaging acquiring a scene in several hundreds of contiguous
spectral bands has opened a new range of relevant applications such as target detection \cite{Manolakis2002} and spectral unmixing \cite{Bioucas2012}. However, while HS sensors provide abundant spectral information, their spatial resolution is generally more limited \cite{Chang2007,Winter2002resolution}. To obtain images with good spectral and
spatial resolutions, the remote sensing community has been devoting increasing research efforts to the problem of fusing HS with MS or PAN images \cite{Cetin2009Merging,Chisense2012,Chen2012super,He2014}. From an application point of view, this problem is also important as motivated by recent national programs, e.g., the Japanese next-generation space-borne hyperspectral image suite (HISUI), which fuses co-registered MS and HS images acquired over the same scene under the same conditions \cite{Yokoya2013}.

The fusion of HS and MS differs from traditional pansharpening since both spatial and spectral information is
contained in multi-band images. Therefore, a lot of pansharpening methods, such as component substitution \cite{Shettigara1992} and relative spectral contribution \cite{Zhou1998wavelet} are inapplicable or inefficient
for the HS/MS fusion problem. Since the fusion problem is generally ill-posed, Bayesian
inference offers a convenient way to regularize the problem by
defining an appropriate generic prior for the scene of interest.
Following this strategy, Gaussian or $\ell_2$-norm priors have been considered to build various
estimators, in the image domain \cite{Hardie2004,Wei2013,Wei2014Bayesian}
or in a transformed domain \cite{Zhang2009}. Recently, the fusion of HS and MS images based on 
spectral unmixing has been explored \cite{Berne2010,Bieniarz2011,Yokoya2012coupled,Yokoya2012Nonlinear}.

Besides, sparse representations have received a considerable interest in recent years exploiting the self-similarity properties of natural images \cite{Shechtman2007,Mairal2007,Mairal2009,Deselaers2010}.
Using this property, a sparse constraint has been proposed in \cite{Yang2010,Zhao2011hyperspectral,Yin2013} to regularize various ill-posed super-resolution and/or fusion problems. The linear decomposition of an image using a few atoms of a redundant dictionary learned from this image (instead of a predefined dictionary, e.g., of wavelets) has recently been used for several problems related to low-level image processing tasks such as denoising \cite{Elad2006} and classification \cite{Ramirez2010}, demonstrating the ability of sparse representations to model natural images. Learning a dictionary from the image of interest is commonly referred to as dictionary learning (DL). Liu \emph{et al.} recently proposed to solve the pansharpening problem based on DL \cite{Liu2012}. DL has also been investigated to restore HS images \cite{Xing2012}. More precisely, a Bayesian scheme was introduced in \cite{Xing2012} to learn a dictionary from an HS image, which imposes a self-consistency of the dictionary by using Beta-Bernoulli processes. This Monte Carlo-based
method provided interesting results at the price of a high computational complexity. Fusing multiple images using a sparse regularization based on the decomposition of these images into high and low frequency
components was considered in \cite{Yin2013}. However, this method developed in \cite{Yin2013} required a training dataset to learn the dictionaries. The references mentioned before proposed to solve the 
corresponding sparse coding problem either by using greedy algorithms such as matching pursuit (MP) and orthogonal MP \cite{Tropp2007} or by relaxing the $\ell_0$-norm to $\ell_1$-norm to take advantage of the last absolute shrinkage and selection operator (LASSO) \cite{Tibshirani1996}.

In this paper, we propose to fuse HS and MS images within a constrained
optimization framework, by incorporating a sparse regularization using dictionaries 
learned from the observed images. Knowing the trained dictionaries and the corresponding
supports of the codes circumvents the difficulties inherent to the sparse coding step. The optimization problem can then be solved by optimizing alternatively with respect to (w.r.t.) the projected target image and the sparse code. The optimization w.r.t. the image is achieved by the split augmented Lagrangian shrinkage
algorithm (SALSA) \cite{Afonso2011}, which is an instance of the alternating direction method
of multipliers (ADMM). By a suitable choice of variable splittings, SALSA enables to 
decompose a huge non-diagonalizable quadratic problem into a sequence of convolutions and
pixel decoupled problems, which can be solved efficiently. The coding step is performed 
using a standard least-square (LS) algorithm which is possible because the supports have been fixed a priori.

The paper is organized as follows. Section \ref{sec:pro} formulates
the fusion problem within a constrained optimization framework. Section \ref{sec:regularizer}
presents the proposed sparse regularization and the method used to learn the dictionaries and the code support. The strategy investigated to solve the resulting optimization problem is detailed in Section \ref{sec:opti}.
Simulation results are presented in Section \ref{sec:simulation} whereas conclusions are reported in Section \ref{sec:conclusions}.

\section{Problem Formulation}
\label{sec:pro}
\subsection{Notations and observation model}

In this paper, we consider the fusion of hyperspectral (HS) and multispectral (MS) images.
The HS image is supposed to be a blurred and down-sampled version of
the target image whereas the MS image is a spectrally degraded version of the target image.
Both images are contaminated by white Gaussian noises.
Instead of resorting to the totally vectorized notations used in \cite{Hardie2004,Zhang2009,Wei2013},
the HS and MS images are reshaped band-by-band to build $\nbbandima \times m$ and $n_{\lambda} \times n$ matrices, respectively, where $\nbbandima$ is the number of HS bands, $n_{\lambda} < \nbbandima$ is the number 
of MS bands, $n$ is the number of pixels in each band of the MS image and $m$ is the number of pixels in each band of the HS image. The resulting observation models associated with the HS and MS images can be written as follows \cite{Hardie2004,Molina1999,Molina2008}

\begin{equation}
\begin{array}{ll}
\label{eq:HS_MS_obs}
\bfY_{\mathrm{H}} =  \MATima \bf{BS} + \bfN_{\mathrm{H}} \\
\bfY_{\mathrm{M}} =  \bfR \MATima + \bfN_{\mathrm{M}}
\end{array}
\end{equation}
where
\begin{itemize}
\item $\MATima = \left[\bsx_1,\ldots,\bsx_{n}\right] \in \mathbb{R}^{\nbbandima \times n}$ is
	the full resolution target image with $\nbbandima$
	bands and $n$ pixels,
\item ${\bfY}_{\mathrm{H}} \in \mathbb{R}^{\nbbandima \times m}$ and ${\bfY}_{\mathrm{M}}\in\mathbb{R}^{n_{\lambda} \times n}$ are the observed HS
and MS images, respectively,
\item $\bfB \in \mathbb{R}^{n \times n}$ is a cyclic convolution operator acting on the bands,
\item ${\bf S}\in\mathbb{R}^{n\times m}$ is a down-sampling matrix (with down-sampling factor denoted as $d$),
\item ${\bfR}\in\mathbb{R}^{n_{\lambda} \times \nbbandima}$ is the spectral
response of the MS sensor,
\item $\bfN_{\mathrm{H}}$ and $\bfN_{\mathrm{M}}$ are the HS and MS noises.
\end{itemize}

Note that $\bfB$ is a sparse symmetric Toeplitz matrix
for a symmetric convolution kernel and $m = n/d^2$, where
$d$ is an integer standing for the downsampling factor. 
In this work, each column of the noise matrices $\bfN_{\mathrm{H}} = \left[\bfn_{\mathrm{H},1}, \ldots \bfn_{\mathrm{H},m}\right]$
and $\bfN_{\mathrm{M}} = \left[\bfn_{\mathrm{M},1},\ldots \bfn_{\mathrm{M},n}\right]$
is assumed to be a band-dependent Gaussian noise vector, i.e., $\bfn_{\mathrm{H},i} \sim
\calN \left(\Vzeros{\nbbandima},\NoiCovMat_{\mathrm{H}}\right)
(i=1,\ldots,m)$ and $\bfn_{\mathrm{M},i} \sim \calN \left(\Vzeros{n_{\lambda}},
\NoiCovMat_{\mathrm{M}}\right) (i=1,\ldots,n)$ where $\Vzeros{a}$ is the $a \times
1$ vector of zeros, $\NoiCovMat_{\mathrm{H}} = \textrm{diag}(\noisevar{\mathrm{H},1},\ldots,\noisevar{\mathrm{H},\nbbandima})
\in \mathbb{R}^{\nbbandima \times \nbbandima}$ and $\NoiCovMat_{\mathrm{M}} = \textrm{diag}(\noisevar{\mathrm{M},1},\ldots,
\noisevar{\mathrm{M},n_{\lambda}}) \in \mathbb{R}^{n_{\lambda} \times n_{\lambda}}$ are diagonal matrices.
Note that the Gaussian noise assumption used in this paper is quite popular in
image processing \cite{Jalobeanu2004,Duijster2009,Xu2011} as it
facilitates the formulation of the likelihood and the associated
optimization algorithms. By denoting the Frobenius norm as $\|{.}\|_F$ ,
the signal to noise ratios (SNRs) of each band in the two images (expressed in decibels)
are defined as
\begin{equation*}
\begin{array}{ll}
\text{SNR}_{\mathrm{H},i} = 10\log \left( \frac{\|{(\MATima \bfB \bfS)_i}\|_F^2}{\noisevar{\mathrm{H},i}} \right), i=1,\ldots,\nbbandima\\
\text{SNR}_{\mathrm{M},j} = 10\log \left( \frac{\|{(\bfR \MATima)_j}\|_F^2}{\noisevar{\mathrm{M},j}} \right), j=1,\ldots,n_{\lambda}.
\end{array}
\end{equation*}

\subsection{Subspace learning}
\label{subsec:Sub_Learn}
The unknown image is $\MATima = \left[\bsx_1,\ldots,\bsx_{n}\right]$ where
$\bsx_i=\left[x_{i,1},x_{i,2},\ldots,x_{i,\nbbandima}\right]^T$
is  the $\nbbandima \times 1$ vector corresponding to the $i$th
spatial location (with $i=1,\ldots,n$). As the bands of the HS data
are generally spectrally dependent, the HS vector $\bsx_i$ usually
lives in a subspace whose dimension is much smaller than the number
of bands $\nbbandima$ \cite{Chang1998,Bioucas2008}, i.e.,
\begin{equation}
\label{eq:subspace}
\bsx_i= \bfH \bsu_i
\end{equation}
where $\bsu_i$ is the projection of the vector $\bsx_i$  onto
the subspace spanned by the columns of $\bfH \in \mathbb{R}^{\nbbandima  \times \wtm_{\lambda}}$
($\bfH$ is an orthogonal matrix ${\bfH}^T {\bfH}=\Id{\wtm_{\lambda}}$).
Using the notation $\bfU = \left[\bsu_1,\ldots,\bsu_{n}\right]$,
we have $\MATima={\bfH}\bfU$ $\left(\bfU \in \mathbb{R}^{\wtm_{\lambda} \times n}\right)$.
Moreover, $ \bfU={\bfH}^T\MATima$ since $\bfH$ is an orthogonal matrix.
In this case, the fusion problem \eqref{eq:HS_MS_obs} can be reformulated
as estimating the unknown matrix $\bfU$ from the following observation equations
\begin{equation}
\begin{array}{ll}
\label{eq:HS_MS_obs2}
\bfY_{\mathrm{H}} =  {\bfH}\bfU \bf{BS} + \bfN_{\mathrm{H}} \\
\bfY_{\mathrm{M}} =  \bfR {\bfH}\bfU+ \bfN_{\mathrm{M}}.
\end{array}
\end{equation}
The dimension of the subspace $\wtm_{\lambda}$ is generally much smaller than
the number of HS bands, i.e., $\wtm_{\lambda} \ll  \nbbandima$. As a consequence,
inferring in the subspace $\mathbb{R}^{\wtm_{\lambda} \times 1}$ greatly decreases
the computational burden of the fusion algorithm. Another motivation for working
in the subspace associated with $\bfU$ is to bypass the possible matrix singularity
caused by the spectral dependency of the HS data. Note that each column of the
orthogonal matrix $\bfH$ can be interpreted as a basis of the subspace of interest.
In this paper, the subspace transform matrix $\bfH$ has been determined from a principal component analysis (PCA)
of the HS data $\bfY_{\mathrm{H}} = \left[\bfy_{\mathrm{H},1},\ldots,\bfy_{\mathrm{H},m}\right]$ (see step 7 of Algorithm \ref{Algo:Alter_Opti}).

\section{Proposed Fusion Rule for multispectral and hyperspectral images}
\label{sec:regularizer}

\subsection{Ill-posed inverse problem}
As shown in \eqref{eq:HS_MS_obs2}, recovering the projected high-spectral and high-spatial resolution image $\bfU$ from the observations $\bfY_{\mathrm{H}}$ and $\bfY_{\mathrm{M}}$ is a linear inverse problem (LIP) \cite{Afonso2011}. In most single-image restoration problem (using either $\bfY_{\mathrm{H}}$ or $\bfY_{\mathrm{M}}$), the inverse problem is ill-posed or under-constrained \cite{Yang2010}, which requires regularization or prior information (in Bayesian terminology).
However, for multi-source image fusion, the inverse problem can be ill-posed or well-posed, depending on the dimension of the subspace and the number of spectral bands. If the matrix $\bf RH$ has full column rank and is well conditioned, which is seldom the case, the estimation of $\bfU$ according to \eqref{eq:HS_MS_obs2} is an over-determined problem instead of an under-determined problem \cite{Germund2008numerical}.
In this case, it is redundant to introduce regularizations. Conversely, if there are fewer MS bands than the subspace dimension $\wtm_{\lambda}$
(e.g., the MS image degrades to a PAN image), the matrix $\bf RH$ cannot have full column rank, which means the fusion problem
is an ill-posed LIP. In this paper, we focus on the under-determined case. Note however that the over-determined problem can be viewed
as a special case with a regularization term set to zero. Another motivation for studying the under-determined problem is that it
includes an archetypal fusion task referred to as pansharpening \cite{Amro2011survey}.

Based on the model \eqref{eq:HS_MS_obs2} and the noise assumption, the distributions of $\bfY_{\mathrm{H}}$ and $\bfY_{\mathrm{M}}$ are
\begin{equation}
\begin{array}{ll}
\label{eq:likelihood}
\bfY_{\mathrm{H}}|\bfU \sim \mathcal{MN}_{\nbbandima,m}({\bfH}\bfU \bfB \bfS, \NoiCovMat_{\mathrm{H}}, \Id{m}), \\
\bfY_{\mathrm{M}}|\bfU \sim \mathcal{MN}_{n_{\lambda},n}(\bfR {\bfH}\bfU, \NoiCovMat_{\mathrm{M}}, \Id{n}).
\end{array}
\end{equation}
where $\mathcal{MN}$ represents the matrix normal distribution. The probability density function
of a matrix normal distribution  ${\cal MN}({\bf M}, \mathbf{\Sigma}_r, \mathbf{\Sigma}_c)$ is defined by
\begin{equation*}
p(\mathbf{X}|\mathbf{M}, \bs{\Sigma}_r, \bs{\Sigma}_c) = \frac{\exp\left( -\frac{1}{2} \, \mathrm{tr}\left[ \bs{\Sigma}_c^{-1} (\mathbf{X} - \mathbf{M})^{T} \bs{\Sigma}_r^{-1} (\mathbf{X} - \mathbf{M}) \right] \right)}{(2\pi)^{np/2} |\bs{\Sigma}_c|^{n/2} |\bs{\Sigma}_r|^{p/2}}
\end{equation*}
where $\bfM$ is the mean, $\bs{\Sigma}_r$ and $\bs{\Sigma}_c$ are two matrices denoted as the row and column covariance matrices.

According to Bayes' theorem and using the fact that the noises ${\bf N}_H$ and ${\bf N}_M$ are independent,
the posterior distribution of $\bfU$ can be inferred as
\begin{equation}
\label{eq:posterior_joint}
  p\left(\bfU|\bfY_{\mathrm{H}},\bfY_{\mathrm{M}}\right) \propto p\left(\bfY_{\mathrm{H}}|\bfU\right)p\left(\bfY_{\mathrm{M}}|\bfU\right)p\left(\bfU\right).
\end{equation}

In this work, we compute the maximum a posteriori (MAP) estimator in an optimization framework 
to solve the fusion problem. Taking the negative logarithm of the posterior distribution,
maximizing the posterior distribution is equivalent to solving the following minimization problem
\begin{equation}
\label{eq:obj_function}
 \min_{\bf U}
          \underbrace{\frac{1}{2}\big\|\NoiCovMat_{\mathrm{H}}^{-\frac{1}{2}}({\bf Y}_{\mathrm{H}}-{\bf HUBS})\big\|_F^2}_{\substack{\text{HS data term}\\ \propto  \ln p({\bf Y}_{\mathrm{H}}|{\bfU})}}  +
          \underbrace{\frac{1}{2}\big\|\NoiCovMat_{\mathrm{M}}^{-\frac{1}{2}}({\bf Y}_{\mathrm{M}}-{\bf RHU})\big\|_F^2}_{\substack{\text{MS data term} \\ \propto  \ln p({\bf Y}_{\mathrm{M}}|{\bfU})}}  +
          \underbrace{\lambda\phi(\bfU)}_{\substack{\text{regularizer}\\ \propto \ln p({\bfU})}}
\end{equation}
where the two first terms are associated with the MS and HS images (data fidelity terms)
and the last term is a penalty ensuring appropriate regularization. Note that $\lambda$ is
a parameter adjusting the importance of regularization w.r.t. the data fidelity terms. It is 
also noteworthy that the MAP estimator is equivalent with the minimum mean square error (MMSE)
estimator when $\phi(\bfU)$ has a quadratic form, which is the case in our approach detailed below.

\subsection{Sparse Regularization}
\label{subsec:Spar_Rep}

Based on the self-similarity property of natural images, modeling images with
a sparse representation has been shown to be very effective in many signal processing
applications \cite{Chen1998}. Generally, an over-complete dictionary with columns
referred to as atoms is proposed as a basis for the image patches. In many applications,
the dictionary $\bfD$ is predefined and can be constructed from wavelets \cite{Mallat1999},
curvelets \cite{Starck2002} or discrete cosine transform (DCT) vectors \cite{AHMED1974}.
However, these bases are not necessarily well matched to natural or remote sensing images
since they do not necessarily adapt to the spatial nature of the observed images. As a consequence,
learning the dictionary from the observed images instead of using predefined bases generally
improves image representation \cite{Elad2006}, which is preferred in most scenarios.
Therefore, an adaptive sparse image-dependent regularization is explored in this paper to solve
the fusion problem of interest. The rationale with adaptive sparse representations
is to learn dictionaries from the data yielding sparse representations thereof. In this case,
the atoms of the dictionary are tuned to the input images, leading to much better results than
predefined dictionaries. More specifically, the goal of sparse regularization is to represent the
patches of the target image as a weighted linear combination of a few elementary basis vectors or atoms,
chosen from a learned overcomplete dictionary. The sparse regularization investigated in this paper
is defined as
\begin{equation}
\label{eq:regul}
\phi(\bfU)=\frac{1}{2}\sum_{i=1}^{\wtm_{\lambda}} \big\|{\bfU_i -\calP\left(\bar\bfD_i \bar\bfA_i\right) \big\|}_F^2,
\end{equation}
where
\begin{itemize}
\item $\bfU_i \in \mathbb{R}^{n}$ is the $i$th band (or row) of $\bfU \in \mathbb{R}^{\wtm_{\lambda} \times n}$, with $i=1,\ldots,\wtm_{\lambda}$,
\item $\calP(\cdot): \mathbb{R}^{n_{\textrm{p}} \times n_{\textrm{pat}}} \mapsto \mathbb{R}^{n \times 1}$ is a linear operator that averages
the overlapping patches\footnote{Note that the overlapping decomposition
adopted here is to prevent the block artifacts \cite{Guleryuz2006}.} of each band,
\item $\bar\bfD_i \in \mathbb{R}^{n_{\textrm{p}} \times n_{\textrm{at}}}$ is the overcomplete dictionary whose columns
are basis elements of size $n_{\textrm{p}}$ (corresponding to the size of a patch),
\item $\bar\bfA_i \in \mathbb{R}^{n_{\textrm{at}} \times n_{\textrm{pat}}}$ is the $i$th band code
($n_{\textrm{at}}$ is the number of atoms and $n_\textrm{pat}$ is the number of patches associated
with the $i$th band).
\end{itemize}

Note that there are $\wtm_{\lambda}$ vectors $\bfU_i \in \mathbb{R}^{n}$ since
the dimension of the hyperspectral subspace in which the observed vectors $\bsx_i$ have been projected
is $\wtm_{\lambda}$. The operation decomposing each band into overlapping patches of
size $\sqrt{n_{\textrm{p}}} \times \sqrt{n_{\textrm{p}}}$ is denoted as
 $\calP^{\ast}(\cdot): \mathbb{R}^{n \times 1} \mapsto \mathbb{R}^{n_{\textrm{p}} \times n_{\textrm{pat}}}$,
 which is the adjoint operation of $\calP(\cdot)$, i.e., $\calP(\calP^{\ast}(\bfX))=\bfX$.

\subsection{Dictionary learning step}
\label{subsec:Dictionary}
The DL strategy advocated in this paper consists of learning the dictionaries $\bar\bfD_i$ and an associated sparse code $\bar\bfA_i$
for each band of a rough estimation of $\bfU$ using the observed HS and MS images. A rough
estimation of $\bfU$, referred as $\tilde{\bfU}$ is constructed using
the MS image $\bfY_{\mathrm{M}}$ and the HS image $\bfY_{\mathrm{H}}$,
following the strategy initially studied in \cite{Hardie2004}. A brief introduction of
this method is given in the Appendix. Note that other estimation methods might also be used to propose a rough estimation of $\bfU$ (see step 1 in Algorithm \ref{Algo:Alter_Opti}). Then each band $\tilde{\bfU}_i$ of $\tilde{\bfU}$ is decomposed into $n_\textrm{pat}$
overlapping patches of size $\sqrt{n_{\mathrm{p}}} \times \sqrt{n_{\mathrm{p}}}$ forming a patch matrix
$\calP^{\ast}(\tilde{\bfU}_i) \in \mathbb{R}^{n_{\mathrm{p}} \times n_\textrm{pat}}$.

Many DL methods have been studied in the recent literature. These methods are for instance based
on K-SVD \cite{Aharon2006}, online dictionary learning (ODL) \cite{Mairal2009} or Bayesian
learning \cite{Xing2012}. In this study, we propose to learn the set $\bar\bfD\triangleq \left[\bar\bfD_1,\ldots,\bar\bfD_{\wtm_{\lambda}}\right]$ of over-complete dictionaries using ODL since it is effective from the computational point of view and has empirically demonstrated to provide more relevant representations. More specifically, the dictionary $\bfD_i$ associated with the band $\bfU_i$
is trained by solving the following optimization problem (see step 3 in Algorithm \ref{Algo:Alter_Opti}).
\begin{equation}
\label{eq:regul_l1}
\left\{\bar\bfD_i, \tilde\bfA_i\right\} = \operatornamewithlimits{argmin}_{\bfD_i, \bfA_i} \frac{1}{2} \left[\big\|{\calP^{\ast}(\tilde{\bfU}_i) -\bfD_i \bfA_i \big\|}_F^2 + \mu \big\|{\bfA_i}\big\|_1\right].
\end{equation}
Then, to provide a more compact representation, we propose to re-estimate the sparse code
\begin{equation}
\label{eq:regul_l0}
\bar\bfA_i = \operatornamewithlimits{argmin}_{\bfA_i} \frac{1}{2} \big\|{\calP^{\ast}(\tilde{\bfU}_i) -\bar\bfD_i \bfA_i \big\|}_F^2,
\ \text{s.t.} \ \big\|{\bfA_i}\big\|_0 \leq K
\end{equation}
where $K$ is a given maximum number of atoms, for each patch of $\bfU_i$. This $\ell_0$-norm constrained regression problem can be addressed using greedy algorithms, e.g., orthogonal matching pursuit (OMP). Generally,
the maximum number of atoms $K$ is set much smaller than the number of atoms in the dictionary, i.e., $K \ll n_\textrm{at}$. The positions of the non-zero elements of the code $\bar\bfA_i$, namely the supports denoted
 $\bar{\bs{\Omega}}_i \triangleq \left\{(j,k)| \bar\bfA_i(j,k) \neq 0 \right\}$, are also identified
(see steps 4 and 5 in Algorithm \ref{Algo:Alter_Opti}).


\subsection{Including the sparse code into the estimation framework}

\begin{figure}[h!]
\centerline{ 
\xymatrix{
  & & & *+<0.05in>+[F-]+{\bar{\bfA}} \ar@{->}[d]& &  & \\
   & *+<0.05in>+[F-]+{\bar{\bfD}} \ar@{->}[rd] &  & \bfA \ar@{->}[ld] &   &\\
  *+<0.05in>+[F-]+ {\NoiCovMat_{\mathrm{H}}} \ar@{->}[rd] &  & \bfU \ar@{->}[ld] \ar@{->}[rd]&  & *+<0.05in>+[F-]+ {\NoiCovMat_{\mathrm{M}}} \ar@{->}[ld] &  &\\
  & \bfY_{\mathrm{H}}& &\bfY_{\mathrm{M}} &  &&
  }
} 
\caption{DAG for the data, parameters and hyperparameters (the fixed parameters appear in boxes).} 
\label{fig:DAG_RCA}
\end{figure}
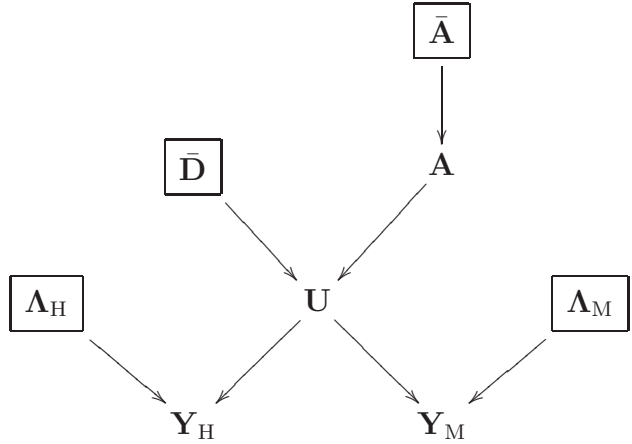

Since the regularization term \eqref{eq:regul} exhibits separable terms w.r.t. each image $\bfU_i$ in band $i$, it can be easily interpreted in a Bayesian framework as the joint prior distribution of the images $\bfU_i$ ($i=1,\ldots,\wtm_{\lambda}$) assumed to be a priori independent, where each marginal prior $p\left(\bfU_i\right)$ is a Gaussian distribution with mean $\calP\left(\bar\bfD_i \bar\bfA_i\right)$. More formally, by denoting $\bar\bfA\triangleq \left[\bar\bfA_1,\ldots,\bar\bfA_{\wtm_{\lambda}}\right]$, the prior distribution for $\bfU$ associated with the regularization \eqref{eq:regul} can be written
\begin{equation}
  p\left(\bfU|\bar\bfD,\bar\bfA\right)= \prod_{i=1}^{\wtm_{\lambda}} p\left(\bfU_i|\bar\bfD_i, \bar\bfA_i\right).
\end{equation}
In a standard approach, the hyperparameters $\bar\bfD$ and $\bar\bfA$ can be a priori fixed, e.g., based on the dictionary learning step detailed in the previous section. However, this choice can drastically impact the accuracy of the representation and therefore the relevance of the regularization term. Inspired by hierarchical models frequently encountered in Bayesian inference \cite{Gelman2013}, we propose to add a second level in the Bayesian paradigm by
fixing the dictionaries $\bar\bfD$ and the set of supports $\bar{\bs{\Omega}} \triangleq \left\{ \bar{\bs{\Omega}}_1,\ldots, \bar{\bs{\Omega}}_{\wtm_{\lambda}}\right\}$, but including the code $\bfA$ within the estimation process. The associated joint prior can be  written as follows, using implicitly $\bar\bfA$ as a hyper-hyperparameter:
\begin{equation}
  p\left(\bfU,\bfA|\bar\bfD,\bar{\bfA}\right)= \prod_{i=1}^{\wtm_{\lambda}} p\left(\bfU_i|\bar\bfD_i, \bfA_i\right)p\left(\bfA_i|\bar{\bfA}_i\right)
\end{equation}
where $\bar{\bs{\Omega}}$ is derived from $\bar{\bfA}$. 
Therefore, the regularization term \eqref{eq:regul} reduces to
\begin{equation}
\label{eq:reg_supp}
\begin{array}{lll}
\phi(\bfU,\bfA)=\frac{1}{2}\sum\limits_{i=1}^{\wtm_{\lambda}} \big\|{\bfU_i-\calP\left(\bar\bfD_i \bfA_i\right)\big\|}_F^2 = \frac{1}{2}\big\|{\bfU- \bar{\bfU}\big\|}_F^2,
\ \text{s.t.}  \ \left\{\bfA_{i,\setminus\bar{\bs{\Omega}}_i} =0\right\}_{i=1}^{\wtm_{\lambda}}
\end{array}
\end{equation}
where $\bar{\bfU}\triangleq\left[\calP\left(\bar\bfD_1 \bfA_1\right),\ldots, \calP\left(\bar\bfD_{\wtm_{\lambda}} \bfA_{\wtm_{\lambda}}\right)\right]$ and $\bfA_{i,\setminus\bar{\bs{\Omega}}_i}= \left\{\bfA_i(j,k) \mid  (j,k) \not\in \bar{\bs{\Omega}}_i \right\}$.
It is worthy to note that i) the regularization term in \eqref{eq:reg_supp} is still separable w.r.t. each band $\bfU_i$, and ii) the optimization of \eqref{eq:reg_supp} w.r.t. $\bfA_i$ reduces to an $\ell_2$-norm optimization task
w.r.t. the non-zero elements in $\bfA_i$, which can be solved easily.
The hierarchical structure of the observed data, parameters and hyperparameters is summarized in Fig. \ref{fig:DAG_RCA}.

Finally, substituting \eqref{eq:reg_supp} into \eqref{eq:obj_function}, the optimization
problem to be solved can be expressed as follows
\begin{multline}
 \operatornamewithlimits{min}_{\bfU, \bfA} L(\bfU, \bfA) \triangleq 
 \frac{1}{2}\big\|\NoiCovMat_{\mathrm{H}}^{-\frac{1}{2}}({\bf Y}_{\mathrm{H}}-{\bf HUBS})\big\|_F^2+
 \frac{1}{2}\big\|\NoiCovMat_{\mathrm{M}}^{-\frac{1}{2}}({\bf Y}_{\mathrm{M}}-{\bf RHU})\big\|_F^2+\frac{\lambda}{2}\big\|{\bfU- \bar{\bfU} \big\|}_F^2,\\
   \text{s.t.}\  \left\{\bfA_{i,\setminus\bar{\bs{\Omega}}_i} =0\right\}_{i=1}^{\wtm_{\lambda}}.
\label{eq:obj_supp}
\end{multline}

Note that the set of constraints  $\left\{\bfA_{i,\setminus\bar{\bs{\Omega}}_i} =0\right\}_{i=1}^{\wtm_{\lambda}}$ could have been removed.
In this case, to ensure sparse representations of $\bfU_i$ ($i=1,\ldots,\wtm_{\lambda}$), sparse constraints on the codes $\bfA_i$ ($i=1,\ldots,\wtm_{\lambda}$),
such as $\left\{\|\bfA_i\|_0<K\right\}_{i=1}^{\wtm_{\lambda}}$ or sparsity promoting penalties, e.g., $\sum\limits_{i=1}^{\wtm_{\lambda}}\|\bfA_i\|_1$ should have been included into the object function \eqref{eq:obj_supp}. This would have resulted in a much more computationally intensive algorithm.
 

\section{Alternate optimization}
\label{sec:opti}


Once $\bar\bfD$, $\bar{\bs{\Omega}}$ and $\bfH$ have been learned from the HS and MS data, the problem
\eqref{eq:obj_supp} reduces to a standard constrained quadratic optimization problem w.r.t. $\bfU$ and $\bfA$. 
However, this problem is difficult to solve due to its large dimension and due
to the fact that the linear operators ${\bf H(\cdot)BD}$ and $\calP(\cdot)$ cannot be easily
diagonalized. To cope with this difficulty, we propose an optimization technique that alternates
optimization w.r.t. $\bfU$ and $\bfA$, which is a simple version of block coordinate descent
algorithm.

The optimization w.r.t. $\bfU$ conditional on $\bfA$ (or equivalent on $\bar{\bfU}$) can be achieved
efficiently with the alternating direction method of multipliers (ADMM) \cite{Boyd2011},
whose convergence has been proved in the convex case. The optimization w.r.t. $\bfA$ with the
support constraint $\bfA_{i,\setminus\bar{\bs{\Omega}}_i} =0$ ($i=1,2,\ldots,\wtm_{\lambda}$) conditional on $\bfU$
is a least squares (LS) regression problem for the non-zero elements of $\bfA$, which can be solved easily. The resulting scheme including
learning $\bar\bfD$, $\bar{\bs{\Omega}}$ and $\bfH$ is detailed in Algo. \ref{Algo:Alter_Opti}.
The alternating ADMM and LS steps are detailed in what follows.

\SetKwComment{tcp}{\%\ }{}
\newcommand{\tcpp}[1]{\tcp{\emph{\small{#1}}}}
\IncMargin{1em}

\begin{algorithm}[h!]
\label{Algo:Alter_Opti}
\KwIn{$\bfY_{\mathrm{H}}$, $\bfY_{\mathrm{M}}$, $\text{SNR}_{\mathrm{H}}$, $\text{SNR}_{\mathrm{M}}$, $\wtm_{\lambda}$ (HS subspace dimension), $\bf R$, $\bfB$, $\bfS$, $K$ }
\tcpp{Propose a rough estimation $\tilde{\bfU}$ of $\bfU$}
Compute $\tilde{\bfU} \triangleq \hat{\bs{\mu}}_{\bfU|\bfY_{\mathrm{M}}}$  following the method in \cite{Hardie2004} (see Appendix)\;
\For{$i=1$ \KwTo $\wtm_{\lambda}$}{
 \tcpp{Online dictionary learning}
${{\bar\bfD_i}}\leftarrow \textsf{ODL} (\tilde{\bfU}_i)$\;
 \tcpp{Sparse coding}
${\bar{\bf A}_i} \leftarrow \textsf{OMP} ({\bar\bfD}_i, \tilde{\bfU}_i, K)$\;
 \tcpp{Identify the supports}
Set ${\bar{\bf \Omega}}_i = \left\{(j,k)|\bar{\bfA}_i(j,k) \neq 0\right\}$\;
}
 \tcpp{Identify the hyperspectral subspace}
${\hat{\bfH}} \leftarrow \textsf{PCA}(\bfY_{\mathrm{H}},\wtm_{\lambda})$\;
 \tcpp{Start alternate optimization}
	   \For{ $t = 1,2, \ldots$ \KwTo stopping rule}{
         \tcpp{Optimize w.r.t. $\bfU$ using SALSA (see Algorithm \ref{Algo:SALSA})}
        ${\hat{\bf U}}^{(t)}  \in \{ {\bf U}|\, L({\bfU}, \hat{\bfA}_{t-1}) \leq L(\hat{\bfU}^{(t-1)} \hat{\bf A}^{(t-1)})\}$\; 
         \tcpp{Optimize w.r.t. $\bfA$ (LS regression)}
        ${\hat{\bf A}}^{(t)}  \in \{ {\bf A}|\, L(\hat{\bfU}^{(t)}, {\bf A}) \leq L(\hat{\bfU}^{(t)}, \hat{\bfA}^{(t-1)})\}$\; 
        }
Set $\hat{\bfX}= \hat{\bf H} \hat{\bfU}$\;
\KwOut{$\hat{\MATima}$ (high resolution HS image)}
\caption{Fusion of HS and MS based on a sparse representation}
\DecMargin{1em}
\end{algorithm}

Note that the objective function is convex w.r.t $\bf (U,A)$ although no strictly. 
In practice, a very simple way to ensure convergence is to add the quadratic terms
$\mu_a\|A\|_F^2 $, with very small $\mu_a$. In this case, the solution of
Algorithm \ref{Algo:Alter_Opti} is unique and the ADMM algorithm converges linearly \cite{Luo1992}.
In practice, we notice that even when $\mu_a$ is zero, the solution of Algorithm \ref{Algo:Alter_Opti}
always converges to a unique point.

\subsection{ADMM Step}
\label{subsec:ADMM}
Recall that the function to be minimized w.r.t. $\bfU$ conditional on $\bfA$ (or $\bar{\bfU}$) is
\begin{equation}
\label{eq:obj_lag}
 \frac{1}{2}\big\|\NoiCovMat_{\mathrm{H}}^{-\frac{1}{2}}({\bf Y}_{\mathrm{H}}-{\bf HUBS})\big\|_F^2+
 \frac{1}{2}\big\|\NoiCovMat_{\mathrm{M}}^{-\frac{1}{2}}({\bf Y}_{\mathrm{M}}-{\bf RHU})\big\|_F^2+
 \frac{\lambda}{2}\big\|{\bfU- \bar{\bfU} \big\|}_F^2.
\end{equation}
By introducing the splittings ${\bf V}_1  = {\bf UB}$, ${\bf V}_2 = {\bfU}$
and ${\bf V}_3 = {\bfU}$ and the respective scaled Lagrange multipliers ${\bf G}_1,{\bf G}_2$ and ${\bf G}_3$,
the augmented Lagrangian associated with the optimization of $\bfU$ can be written as
\begin{eqnarray*}
\label{eq:COST_LAG}
\lefteqn{L({\bfU}, {\bf V}_1, {\bf V}_2, {\bf V}_3, {\bf G}_1,{\bf G}_2,{\bf G}_3) =}  \\
  &&  \frac{1}{2}\big\|\NoiCovMat_{\mathrm{H}}^{-\frac{1}{2}} ({\bf Y}_{\mathrm{H}}-{\bf HV}_1{\bf S})\big\|_F^2  + \
                  \frac{\mu}{2}\big\|{\bf UB}-{\bf V}_1-{\bf G}_1\big\|_F^2 +\\
  &&  \frac{1}{2}\big\|\NoiCovMat_{\mathrm{M}}^{-\frac{1}{2}} ({\bf Y}_{\mathrm{M}}-{\bf RHV}_2)\big\|_F^2  +
                  \frac{\mu}{2}\big\|{\bfU}-{\bf V}_2-{\bf G}_2\big\|_F^2 + \\
  &&  \frac{1}{2}\big\|\bar{\bfU}-{\bf V}_3\big\|_F^2
  +
                  \frac{\mu}{2}\big\|{\bfU}-{\bf V}_3-{\bf G}_3\big\|_F^2.
\end{eqnarray*}

The updates of $\bfU,\bfV_1,\bfV_2,\bfV_3, \bfG_1,\bfG_2$ and $\bfG_3$ are achieved with
an optimization tool named split augmented Lagrangian shrinkage algorithm (SALSA) \cite{Afonso2010,Afonso2011},
which is an instance of the ADMM algorithm with guaranteed convergence.
The SALSA scheme is summarized in Algorithm \ref{Algo:SALSA}. Note that the optimization w.r.t. to
$\bfU$ (step 5) can be efficiently solved in the Fourier domain.

\IncMargin{1em}
\begin{algorithm}[h!]
\label{Algo:SALSA}
\KwIn{$\hat{\bfU}^{(t)}$, $\bar{\bfD}$, $\hat{\bfA}^{(t)}$, $\bfY_{\mathrm{H}}$, $\bfY_{\mathrm{M}}$, $\text{SNR}_{\mathrm{H}}$, $\text{SNR}_{\mathrm{M}}$, $\bfH$, $\bfR$, $\bfB$, $\bfS$, $\lambda$ and $\mu$ (SALSA parameter)}
Set $\bar{\bfU} =\left[\calP\left(\bar\bfD_1 \hat{\bfA}_1^{(t)}\right),\ldots, \calP\left(\bar\bfD_{\wtm_{\lambda}} \hat{\bfA}^{(t)}_{\wtm_{\lambda}}\right)\right]$\;
Set $\boldsymbol{\delta}  \in \{0,1\}^n$  such that $\boldsymbol{\delta}(i) = \left\{
				\begin{array}{ll}
				1  & \textrm{ if pixel $i$ is sampled,}\\
				0  & \textrm{ otherwise;}
				\end{array}
				\right. 		$\\
{\bf Initialization}: $\bfV_1^{(0)}$,$\bfV_2^{(0)}$,$\bfV_3^{(0)}$,
${\bf G}_1^{(0)}$, ${\bf G}_2^{(0)}$,${\bf G}_3^{(0)}$\;

\For{ $k = 0$  \KwTo  $n_{\mathrm{it}}$}
{\tcpp{Optimize w.r.t $\bfU$}
${\hat{\bfU}^{(t,k+1)}}\leftarrow \left[({\bf V}_1^{(k)} + {\bf G}_1^{(k)}){\bf B}^T + (\bfV_2^{(k)} + {\bfG}_2^{(k)}) +
        ({\bf V}_3^{(k)} + {\bf G}_3^{(k)})\right]\left({{\bf BB}^T + 2{\bf I}}\right)^{-1}$\;
\tcpp{Optimize w.r.t ${\bf V}_1$}
$\bs{\nu}_1\leftarrow (\hat{\bfU}^{(t,k+1)} \bfB - \bfG_1^{(k)}) $\;
\tcpp{Optimize $\bfV_1$ (according to down-sampling)}
${\bf V}^{(k+1)}_1(:,\boldsymbol{\delta}) \leftarrow  \left({\bfH}^T \NoiCovMat_{\mathrm{H}}^{-1} {\bfH} + \mu{\bf I}\right)^{-1}({\bfH}^T \NoiCovMat_{\mathrm{H}}^{-1}{\bf Y}_{\mathrm{H}} + \bs{\nu}_1(:,\boldsymbol{\delta}))$\;
${\bf V}^{(k+1)}_1(:,1-\boldsymbol{\delta})\leftarrow  \bs{\nu}_1(:,{1 - \boldsymbol{\delta}})$ \;
$\bs{\nu}_2 \leftarrow  (\hat{\bfU}^{(t,k+1)} - \bfG^{(k)}_2 $)\;
${\bfV}^{(k+1)}_2\leftarrow  \left({\bfH}^T{\bf R}^T \NoiCovMat_{\mathrm{M}}^{-1}{\bf RH} + \mu{\bf I}\right)^{-1}({\bfH}^T{\bf R}^T\NoiCovMat_{\mathrm{M}}^{-1}{\bf Y}_{\mathrm{M}} + \mu\bs{\nu}_2 )$\;
\tcpp{Optimize w.r.t ${\bf V}_3$}
$\bs{\nu}_3\leftarrow  (\hat{\bfU}^{(t,k+1)} - \bfG^{(k)}_3) $\;
${\bf V}^{(k+1)}_3\leftarrow  \left(\lambda  +\mu \right)^{-1}\left(\lambda \bar{\bfU}+\mu\bs{\nu}_3\right)$\;
\tcpp{Update Lagrange multipliers}
${\bf G}^{(k+1)}_1 \leftarrow  (-\bs{\nu}_1 + {\bf V}^{(k+1)}_1)$\;
${\bf G}^{(k+1)}_2 \leftarrow  (-\bs{\nu}_2 + {\bf V}^{(k+1)}_2)$\;
${\bf G}^{(k+1)}_3 \leftarrow  (-\bs{\nu}_3 + {\bf V}^{(k+1)}_3)$\; }
Set $\hat{\bfU}^{(t+1)} = \hat{\bfU}^{(t,n_{\mathrm{it}})}$\;
\KwOut{$\hat{\bfU}^{(t+1)}$}
\caption{SALSA sub-iterations}
\end{algorithm}
\DecMargin{1em}

\subsection{Patchwise Sparse Coding}
The optimization w.r.t. $\bfA$ conditional on $\bfU$ can be formulated as
\begin{equation}
\label{eq:code}
\hat{\bfA}_i = \argmin_{\bfA_i} \big\| {\bfU}_i - \calP (\bar\bfD_i \bfA_i)\big\|_F^2,
\textrm{ s.t. }\bfA_{i,\setminus\bar{\bs{\Omega}}_i} =0, \ i=1,\ldots,\wtm_{\lambda}.
\end{equation}

Since the operator $\calP(\cdot)$ is a linear mapping from patches to images and $\calP\left(\calP^{\ast}\left(\bfX\right)\right) = \bfX$, the problem \eqref{eq:code} can be rewritten as
\begin{equation}
\label{eq:code_pat_1}
\hat{\bfA}_i = \argmin_{\bfA_i} \big\| \calP \left( \calP^{\ast}({\bfU}_i) - \bar\bfD_i \bfA_i \right)\big\|_F^2,
\textrm{ s.t. }\bfA_{i,\setminus\bar{\bs{\Omega}}_i} =0, \ i=1,\ldots,\wtm_{\lambda}.
\end{equation}

Furthermore, as the adjoint operator has the property 
$\calP^{\ast}\left(\calP(\cdot)\right)\approx c \cal I(\cdot)$ (where $c$ is 
a constant and $\cal I$ is the identity operator), the sub-optimal solution of problem \eqref{eq:code_pat_1} can be approximated by solving
\begin{equation}
\label{eq:code_pat}
\hat{\bfA}_i = \argmin_{\bfA_i} \big\|   \calP^{\ast}({\bfU}_i) - \bar\bfD_i \bfA_i \big\|_F^2,
\textrm{ s.t. }\bfA_{i,\setminus\bar{\bs{\Omega}}_i} =0, \ i=1,\ldots,\wtm_{\lambda}.
\end{equation}

Tackling the support constraint consists of only updating  the non-zero elements of each column of $\bfA_i$. Denote the $j$th vectorized column of
$\calP^{\ast}({\bfU}_i)$ as $\bfp_{i,j}$,
the vector composed of the $K$ non-zero elements of the $j$th column of $\bfA_i$ as $\bfa_{\bar{\bs{\Omega}}_i^j}$,
and the corresponding columns of $\bar{\bfD}_i$ 
as $\bar\bfD_{\bar{\bs{\Omega}}_i^j}$. Then the $\wtm_{\lambda}$ problems
in \eqref{eq:code_pat} reduce to $\wtm_{\lambda} \times n_\textrm{pat}$ sub-problems
\begin{equation}
\label{eq:code_obj_ls}
\hat{\bfa}_{\bar{\bs{\Omega}}_i^j}
=\argmin_{\bfa_{\bar{\bs{\Omega}}_i^j}} \big\|{\bfp_{i,j}- \bar\bfD_{\bar{\bs{\Omega}}_i^j} \bfa_{\bar{\bs{\Omega}}_i^j}\big\|}_F^2, \ i=1,\ldots,\wtm_{\lambda},\
j=1,\ldots,n_\textrm{pat}
\end{equation}
whose solutions $\hat{\bfa}_{\bar{\bs{\Omega}}_i^j} = (\bar\bfD_{\bar{\bs{\Omega}}_i,{j}}^T \bar\bfD_{\bar{\bs{\Omega}}_i^{j}})^{-1} {\bar\bfD^T_{\bar{\bs{\Omega}}_i^{j}}} \bfp_{i,j}$ can be explicitly computed in parallel.
The corresponding patch estimate is $\hat{\bfp}_{i,j}\triangleq\bfT_{i,j} \bfp_{i,j}$, with $\bfT_{i,j}=\bar\bfD_{\bar{\bs{\Omega}}_i^{j}}(\bar\bfD^T_{\bar{\bs{\Omega}}_i^{j}} \bar\bfD_{\bar{\bs{\Omega}}_i^{j}})^{-1} {\bar\bfD^T_{\bar{\bs{\Omega}}_i^{j}}}$. These patches are used to build $\bar{\bfU}$ (i.e., equivalently, $\calP\left(\bar\bfD_i\bfA_i\right)$) required in the optimization w.r.t. $\bfU$ (see section~\ref{subsec:ADMM}). Note that  $\bfT_{i,j}$ is a projection operator, and hence is
symmetric ($\bfT_{i,j}^T=\bfT_{i,j}$) and idempotent ($\bfT_{i,j}^2=\bfT_{i,j}$). Note also that $\bfT_{i,j}$ needs to be calculated only once, given the learned dictionaries and associated supports.

\subsection{Complexity Analysis}
The SALSA algorithm has a complexity of the order $\mathcal{O}\left(n_{\mathrm{it}}\wtm_{\lambda}n \log \left(\wtm_{\lambda}n\right)\right)$
\cite{Afonso2011}, where $n_{\mathrm{it}}$ is the number of SALSA iterations.
The computational complexity of the patchwise sparse coding is
$\mathcal{O}\left(K n_{\mathrm{p}} n_\textrm{pat}\wtm_{\lambda}\right)$.
Conducting the fusion in a subspace of dimension $\wtm_{\lambda}$ instead
of working with the initial space of dimension $\nbbandima$
greatly decreases the complexity of both SALSA and sparse coding steps.

\section{Simulation Results on Synthetic Data}
\label{sec:simulation}
\begin{figure}[h!]
\centering
    \subfigure{
    \label{fig:subfig:Ref}
    \includegraphics[width=0.2\textwidth]{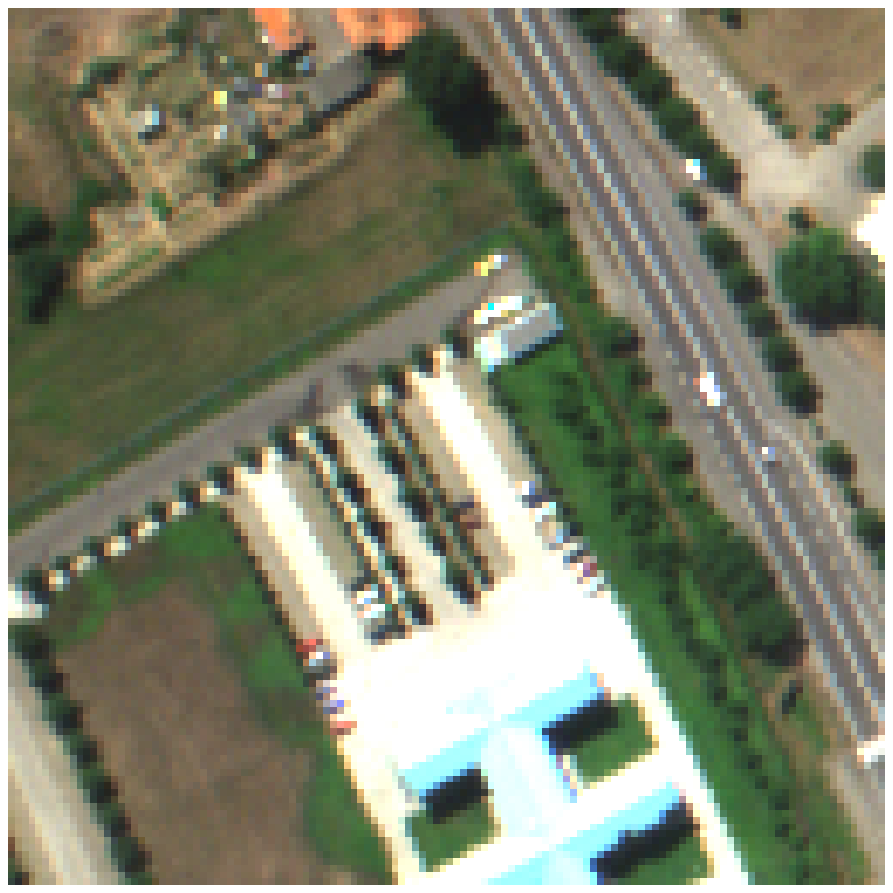}}
    \hspace{0in}
    \subfigure{
    \label{fig:subfig:HS}
    \includegraphics[width=0.2\textwidth]{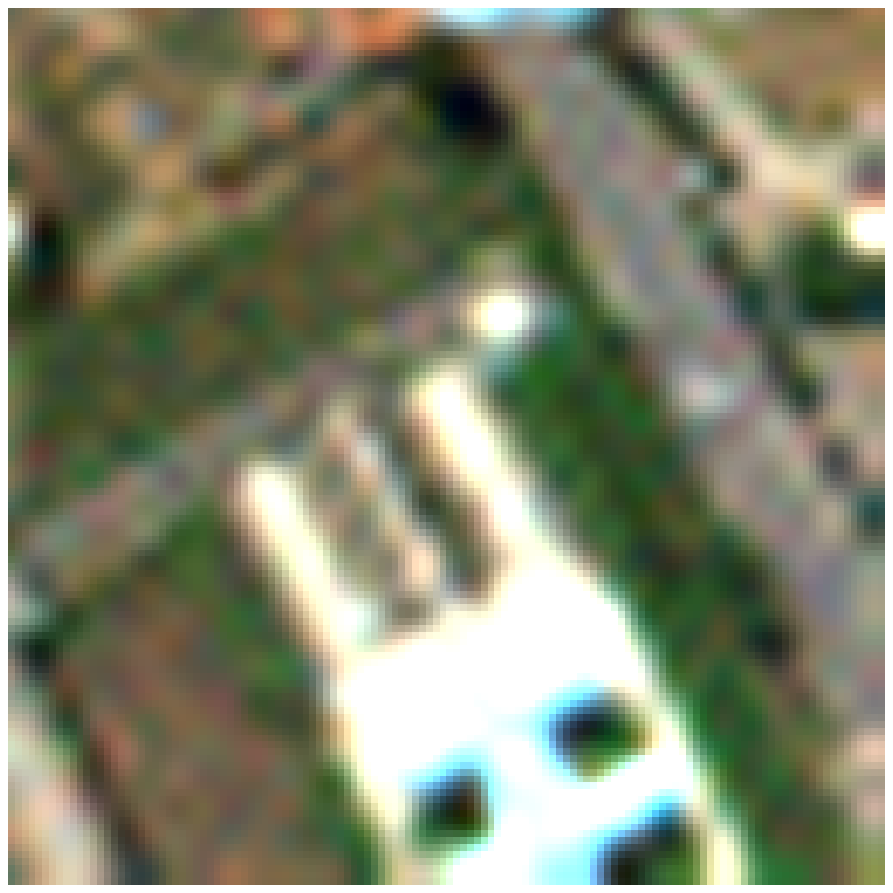}}
    \hspace{0in}
    \subfigure{
    \label{fig:subfig:MS}
    \includegraphics[width=0.2\textwidth]{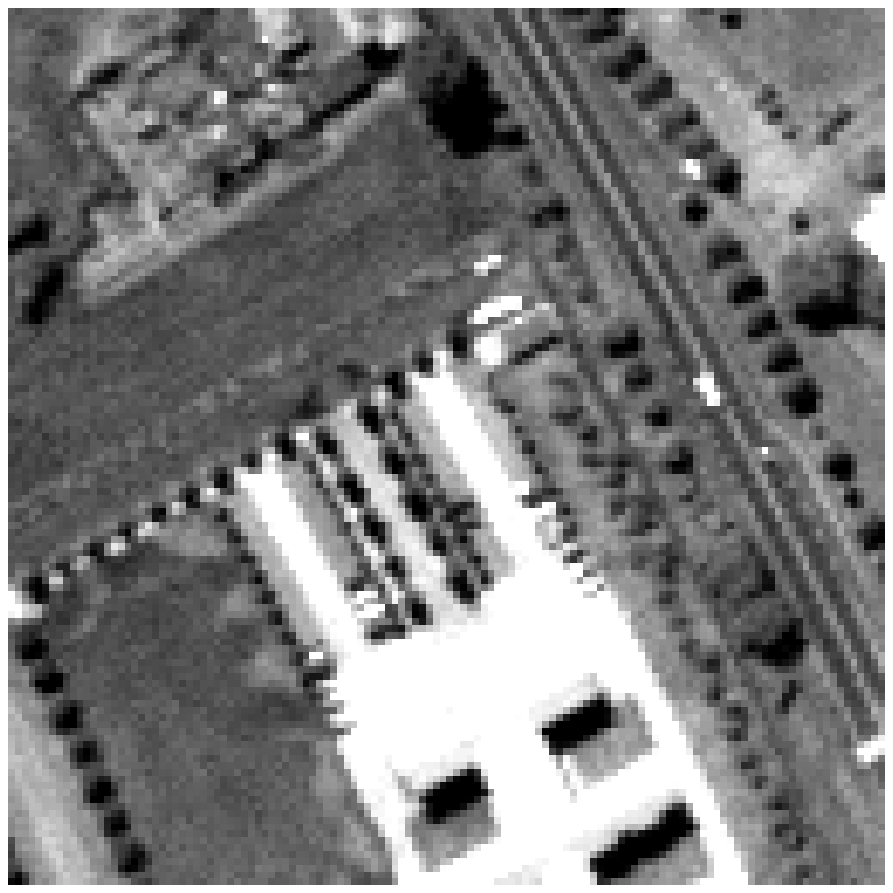}}\\
    \caption{(left) Reference image. (middle) HS Image. (right) MS Image.}
    \label{fig:O_H_M}
\end{figure}

This section studies the performance of the proposed sparse
representation-based fusion algorithm. The reference image
considered here as the high spectral and high spectral image
is a $128 \times 128 \times 93$ HS image with spatial
resolution of $1.3$m acquired by the reflective optics system
imaging spectrometer (ROSIS) optical sensor
over the urban area of the University of Pavia, Italy. The flight was
operated by the Deutsches Zentrum f\"{u}r Luft- und Raumfahrt (DLR, the
German Aerospace Agency) in the framework of the
HySens project, managed and sponsored by the European
Union. This image was initially composed of $115$ bands
that have been reduced to $93$ bands after removing the water
vapor absorption bands (with spectral range from 0.43 to 0.86μm).
It has received a lot of attention in the remote sensing literature
\cite{Plaza2009,Tarabalka2010,Li2013}.  A composite color image, formed by selecting the red, green and blue bands of the reference image is shown in the left of Fig. \ref{fig:O_H_M}.

\subsection{Simulation Scenario}
\label{sec:Scenario}
\begin{figure}[h!]
\centering
	\includegraphics[width=0.5\textwidth]{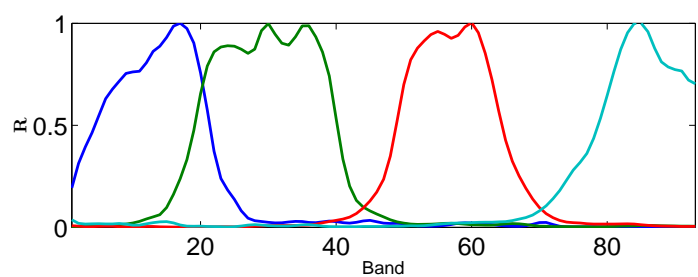}
	\caption{IKONOS-like spectral responses.}
\label{fig:F2}
\end{figure}

We propose to reconstruct the reference HS image
from two lower resolved images. A high-spectral low-spatial resolution HS image
has been constructed by applying a $5 \times 5$ exponential degrading spatial filter on each band of the
reference image and down-sampling every $4$ pixels in both horizontal and vertical
directions. In a second step, we have generated a $4$-band MS image by filtering the
reference image with the IKONOS-like reflectance spectral responses depicted in Fig. \ref{fig:F2}.
The HS and MS images are both perturbed by zero-mean additive Gaussian noises.
Our simulations have been conducted with SNR$_{1,\cdot}=35$dB for the first 127 bands and
SNR$_{1,\cdot}=30$dB for the remaining 50 bands of the HS image. For the MS image, SNR$_{2,\cdot}$
is 30dB for all bands. The noise-contaminated HS and MS images are depicted
in the middle and right of Fig. \ref{fig:O_H_M} (the HS image has been
interpolated for better visualization).

\subsection{Learning the Subspace, the Dictionaries and the Code Supports}
\label{subsec:prepare}
\subsubsection{Subspace}
\label{subsubsec:subspace}
\begin{figure}[h!]
\centering
\includegraphics[width=0.5\textwidth]{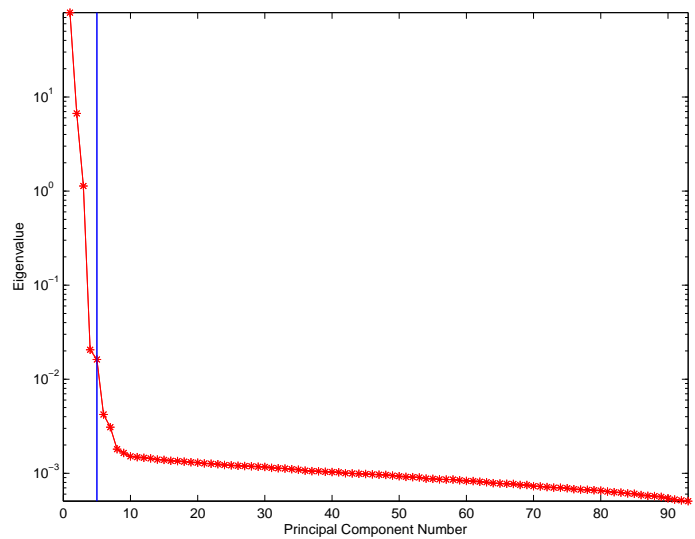}
\caption{Eigenvalues of $\bs{\Upsilon}$ for the Pavia HS image.}
\label{fig:Eigen_value}
\end{figure}

To learn the transform matrix $\bfH$, we propose to use the principal component analysis (PCA)
as in \cite{Wei2014Bayesian}. Note that PCA is a classical dimensionality reduction technique used in
hyperspectral imagery. 
The empirical covariance matrix $\bs{\Upsilon}$ of the HS pixel vectors is diagonalized leading to
\begin{equation}
\mathbf{W}^{T} \bs{\Upsilon} \mathbf{W} = \bs{\Gamma}
\end{equation}
where $\mathbf{W}$ is an $\nbbandima \times \nbbandima$ unitary
matrix ($\mathbf{W}^T=\mathbf{W}^{-1}$) and $\bs{\Gamma}$ is a
diagonal matrix whose diagonal elements are the ordered eigenvalues
of $\bs{\Upsilon}$ denoted as $d_1 \ge d_2 \ge ... \ge
d_{\nbbandima}$. The top $\wtm_{\lambda}$ components are selected
and the matrix $\bfH$ is then constructed as the eigenvectors
associated with the $\wtm_\lambda$ largest eigenvalues of
$\bs{\Upsilon}$.  As an illustration, the eigenvalues of
the empirical covariance matrix $\bs{\Upsilon}$ for the
Pavia image are displayed in Fig. \ref{fig:Eigen_value}. For
this example, the $\wtm_{\lambda}= 5$ eigenvectors contain $99.9 \%$
of the information.

\subsubsection{Dictionaries}
\label{subsubsec:dic_learn}
As explained before, the target high resolution image is assumed to live in
the lower dimensional subspace. Firstly, a rough estimation of the projected image
is obtained with the method proposed in \cite{Hardie2004}. In a second step,
$\wtm_\lambda=5$ dictionaries are learned from the rough estimation of the projected
image using the ODL method.

As $n_{\mathrm{at}} \gg n_{\mathrm{p}}$, the dictionary is over-complete. There is no unique rule to select the
dictionary size $n_{\mathrm{p}}$ and the number of atoms $n_{\mathrm{at}}$. However, two limiting cases can be identified:
\begin{itemize}
\item The patch reduces to a single pixel, which means $n_{\mathrm{p}}=1$. In this case, the sparsity is
	 not necessary to be introduced since only one $1$D dictionary atom (which is a constant) is enough to represent any target patch.
\item The patch is as large as the whole image, which means only one atom is needed to represent the image. In this case, the atom is too ``specialized'' to describe any other image.
\end{itemize}

More generally, the smaller the patches, the more objects the atoms can approximate. However, too small patches are not efficient to properly capture the textures, edges, etc.
With larger patch size, a larger number of atoms are required to guarantee the over-completeness
(which requires larger computation cost). In general, the size of patches is selected empirically.
For the ODL algorithm used in this study, the size has been fixed to $n_{\mathrm{p}}=6 \times 6$
and the number of atoms is $n_{\mathrm{at}}=256$. The learned dictionaries for the first three bands of $\tilde{\bfU}$ are
displayed in the left column of Fig. \ref{fig:Dic_LD}. This figure shows that the spatial properties of
the target image have been captured by the atoms of the dictionaries.

\subsubsection{Code Supports}
\begin{figure}[h!]
\centering
\subfigure[Dictionary for band 1]{
\includegraphics[width=0.4\textwidth]{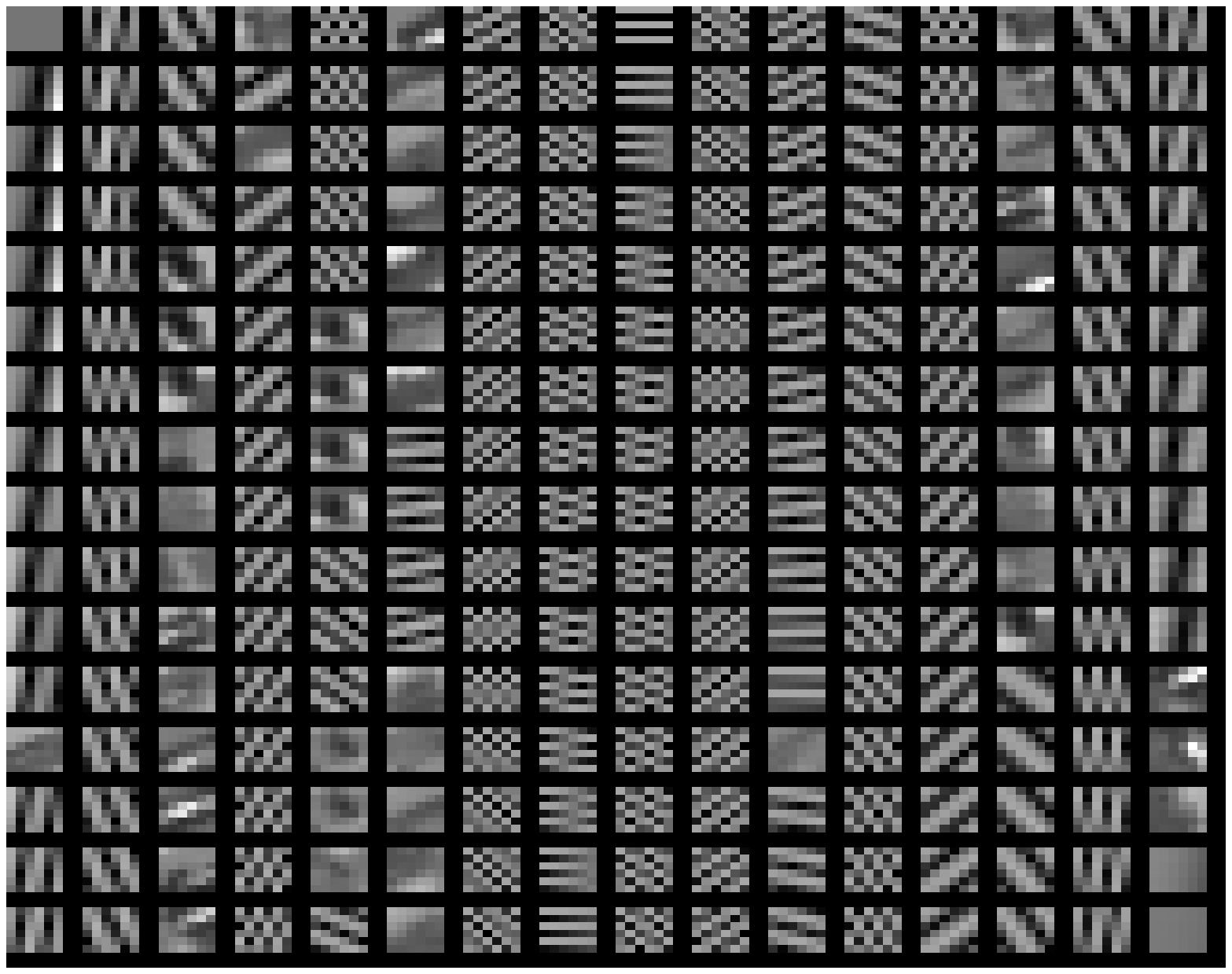}}
\subfigure[Support for band 1 (for some patches)]{
\includegraphics[width=0.4\textwidth]{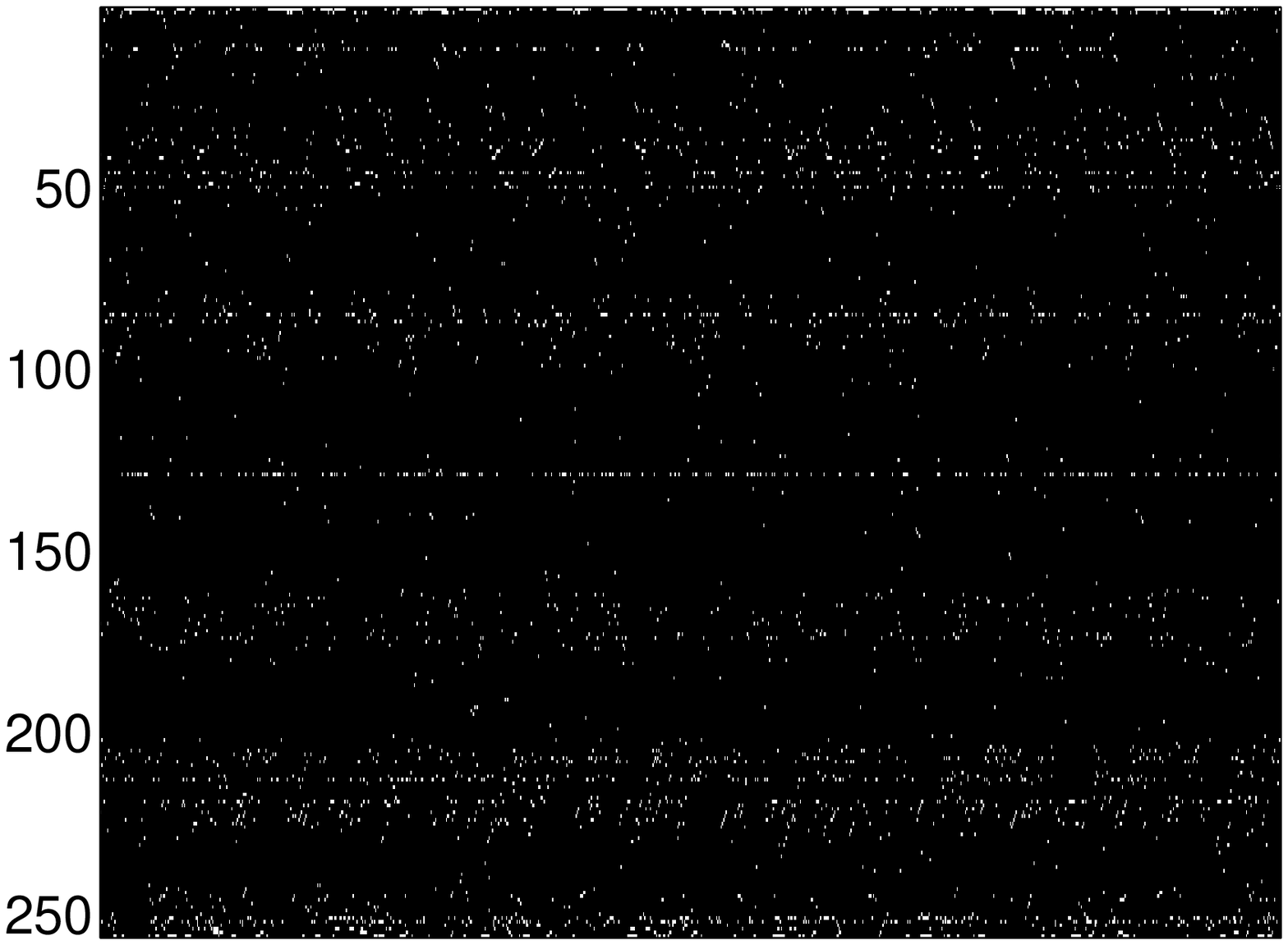}}
\subfigure[Dictionary for band 2]{
\includegraphics[width=0.4\textwidth]{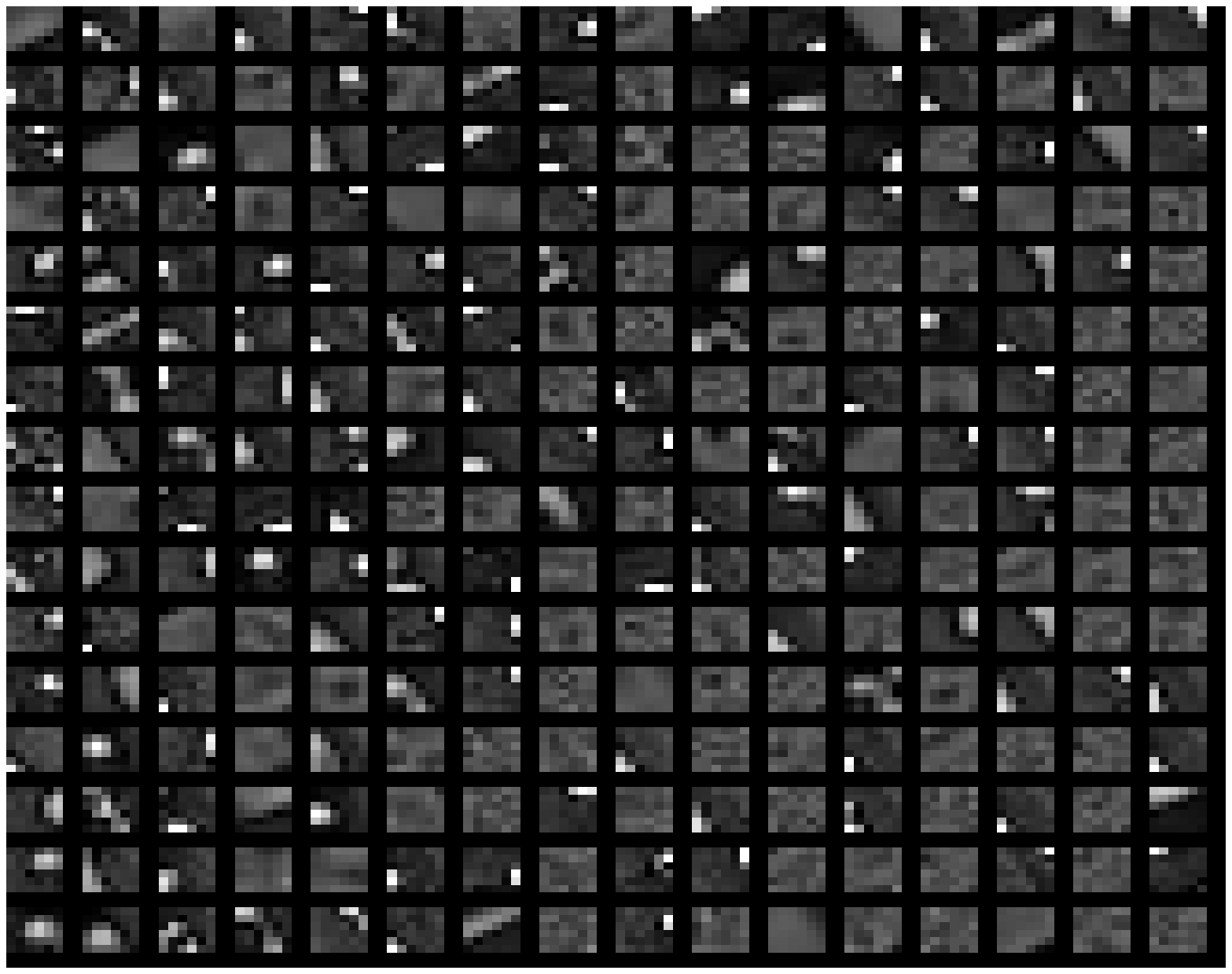}}
\subfigure[Support for band 2 (for some patches)]{
\includegraphics[width=0.4\textwidth]{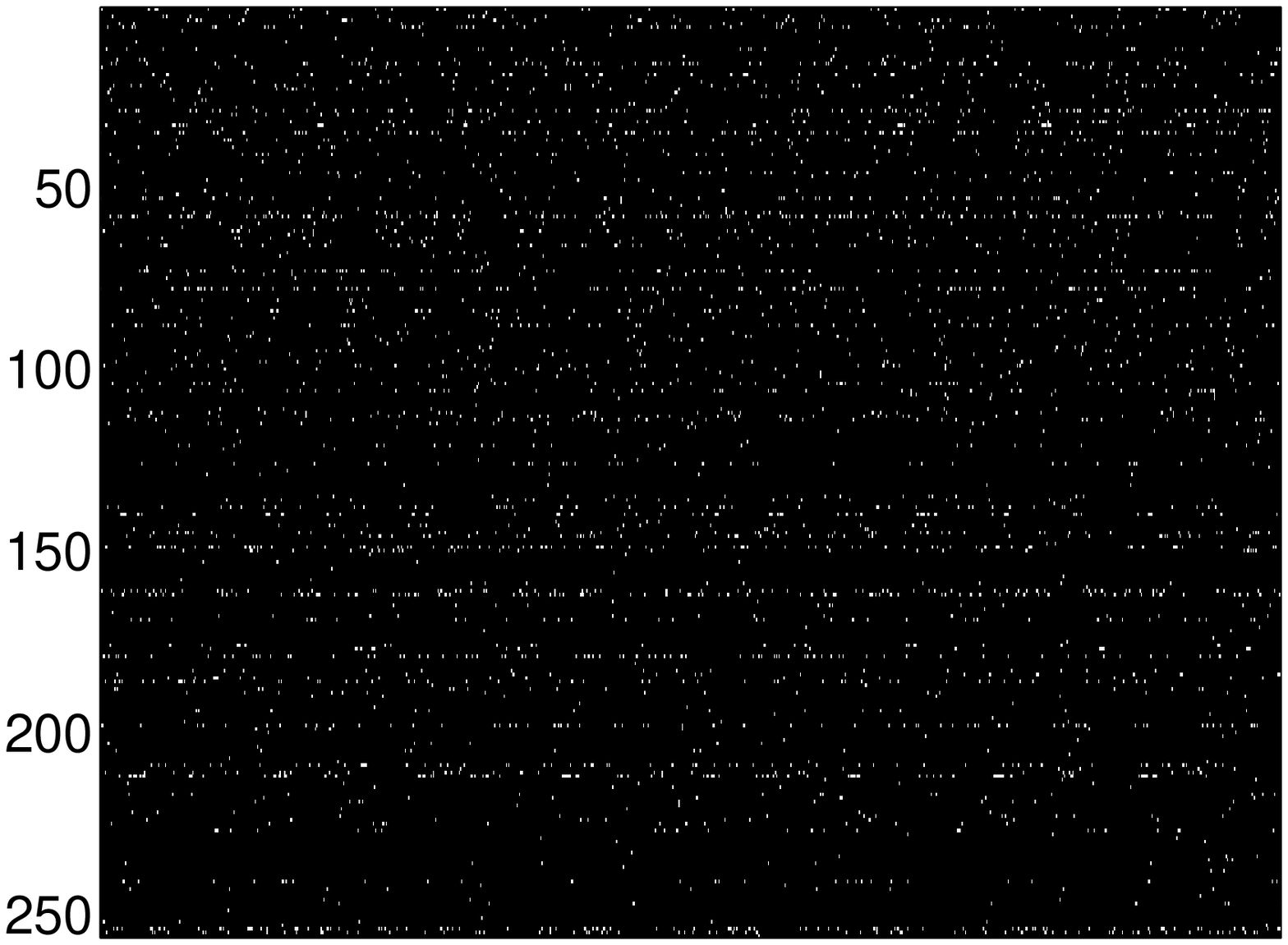}}
\subfigure[Dictionary for band 3]{
\includegraphics[width=0.4\textwidth]{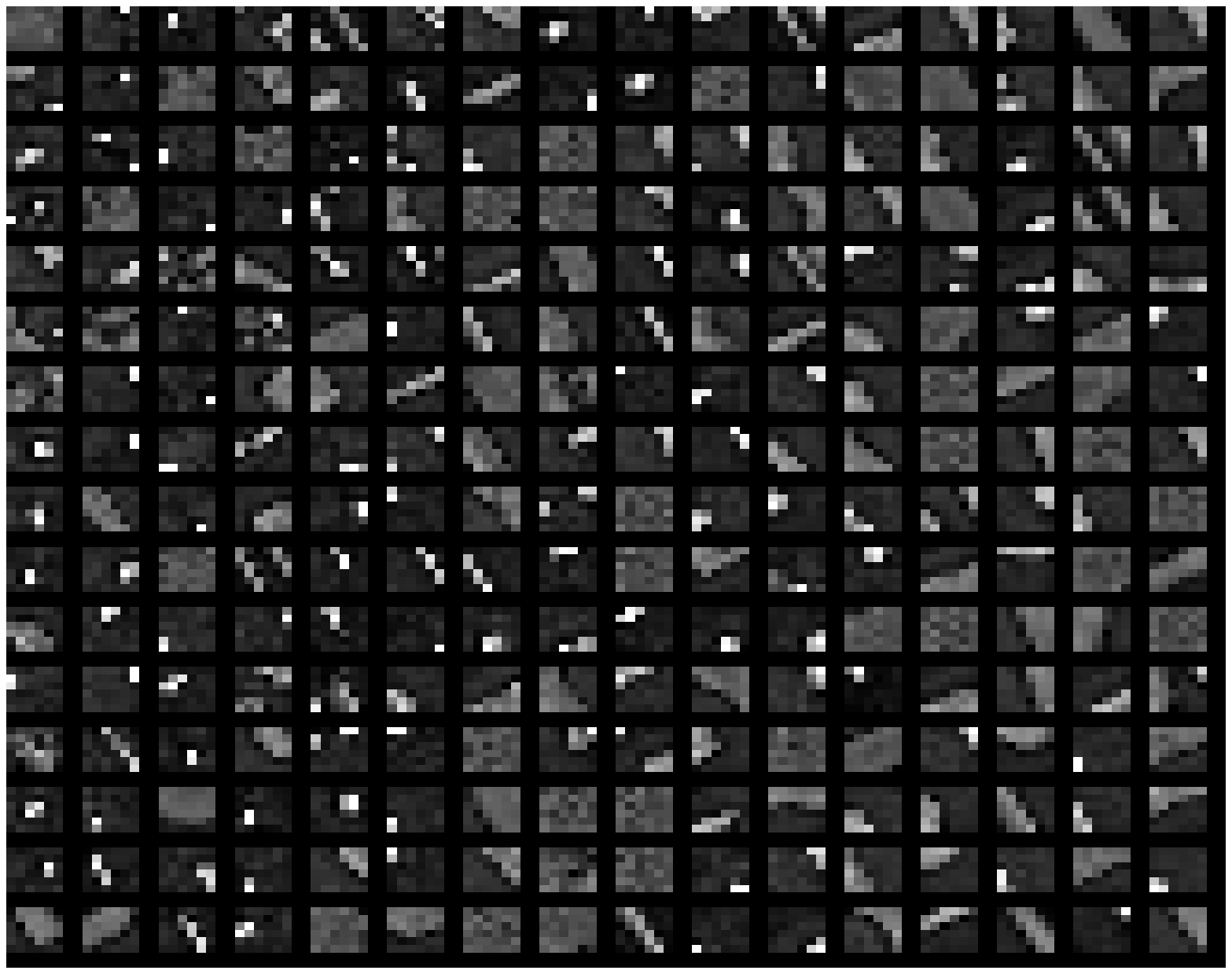}}
\subfigure[Support for band 3 (for some patches)]{
\includegraphics[width=0.4\textwidth]{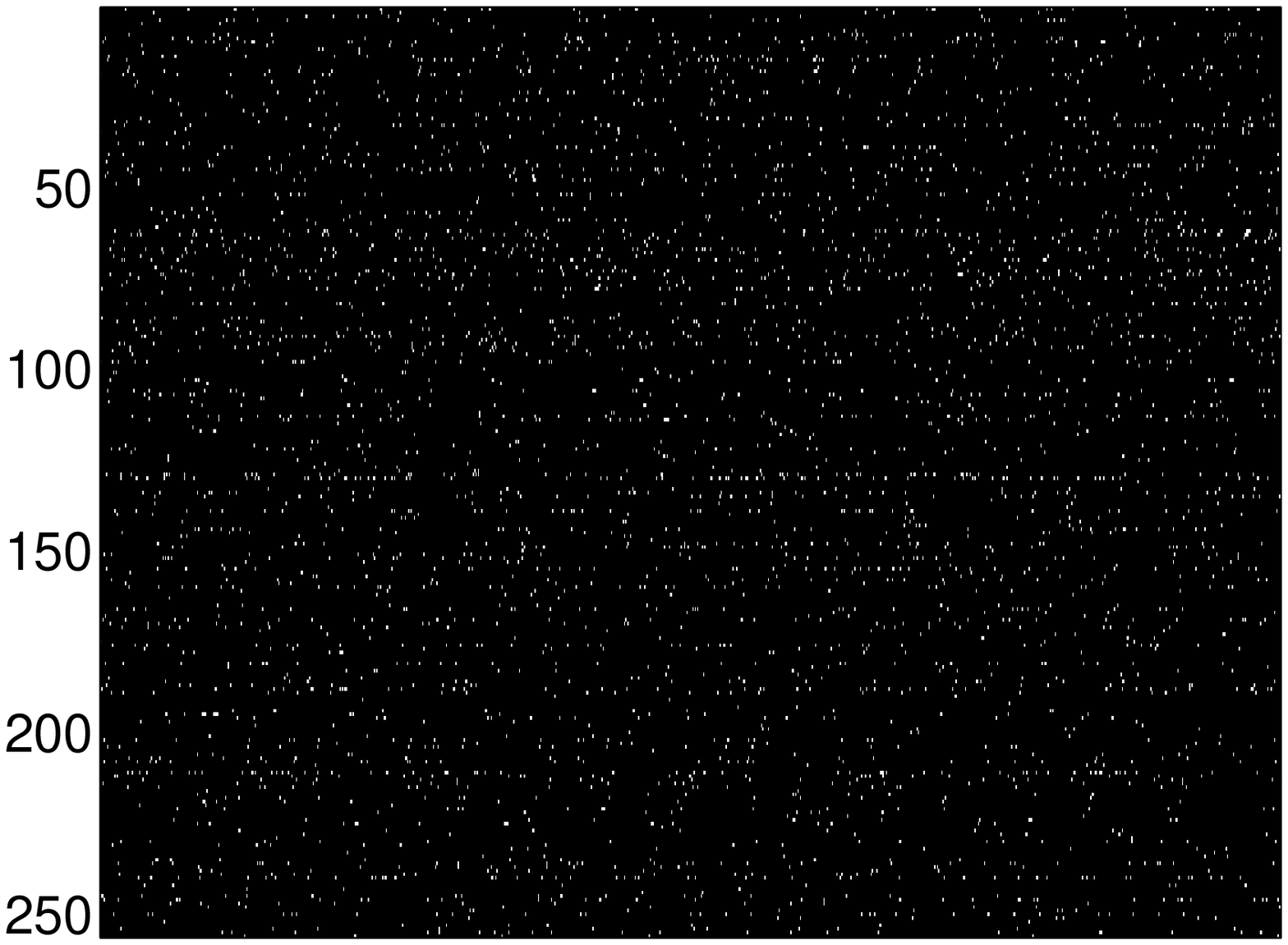}}
\caption{Learned dictionaries (left) and example of corresponding supports (right).}
\label{fig:Dic_LD}
\end{figure}

Based on the dictionaries learned following the strategy presented in Section \ref{subsubsec:dic_learn},
re-estimation of the code is achieved by solving \eqref{eq:regul_l0} with OMP. Note that the target sparsity $K$ represents the maximum number of atoms used to represent one patch, which also determines the number of non-zeros elecments of $\bfA$ estimated jointly with the projected image $\bfU$. If $K$ is too large, the optimization w.r.t. $\bfU$ and $\bfA$ leads to over-fitting, which means there are too many parameters to estimate while the sample size is too small. The training supports for the first three bands are displayed in the right column of Fig. \ref{fig:Dic_LD}. The number of rows is $256$, which represents the number of atoms in each dictionary $\bar\bfD_i$ ($i=1,\ldots,\wtm_{\lambda}$). The white dots in the $j$th column indicate which atoms are used for reconstructing the $j$th patch ($j=1,\ldots,n_\textrm{pat}$). The sparsity is clearly illustrated in this figure. Note that some atoms are frequently used whereas some others are not. The most popular atoms represent spatial details that are quite common in images. The other atoms represent details that are characteristics of specific patches.

\subsection{Fusion Quality Metrics}
To evaluate the quality of the proposed fusion strategy, several image quality measures have been employed. Referring to \cite{Zhang2009}, we propose to use RMSE, SAM, UIQI, ERGAS and DD as defined below.
\subsubsection{RMSE}
The root mean square error (RMSE) is a similarity measure between the target image $\MATima$ and fused image $\hat\MATima$ defined as
\begin{equation*}
\textrm{RMSE}(\MATima,\hat{\MATima})=\frac{1}{n \nbbandima}\|\MATima-\hat{\MATima}\|_F^2.
\label{eq:SNR}
\end{equation*}
The smaller RMSE, the better the fusion quality.

\subsubsection{SAM}
The spectral angle mapper (SAM) measures the spectral distortion between the actual and estimated images.
The SAM of two spectral vectors $\bsx_n$ and $\hat{\bsx}_n$ is defined as
\begin{equation*}
\textrm{SAM}(\bsx_n,\hat{\bsx}_n)=\textrm{arccos} \left(\frac{\langle\bsx_n,\hat{\bsx}_n\rangle}{ \|\bsx_n\|_2\|\hat{\bsx}_n\|_2}\right).
\label{eq:SAM}
\end{equation*}
The overall SAM is finally obtained by averaging the SAMs computed from all image pixels.
Note that the value of SAM is expressed in degrees and thus belongs to $(-90,90]$.
The smaller the absolute value of SAM, the less important the spectral distortion.

\subsubsection{UIQI}
The universal image quality index (UIQI) was proposed in \cite{Wang2002} for evaluating the similarity
between two single band images. It is related to the correlation, luminance distortion and contrast
distortion of the estimated image w.r.t. the reference image. The UIQI between two single-band images $\bfa=[a_1,a_2,\ldots,a_N]$
and $\hat{\bfa}=[\hat{a}_1,\hat{a}_2,\ldots,\hat{a}_N]$ is defined as
\begin{equation*}
\textrm{UIQI}(\bfa,\hat{\bfa}) =
\frac{4\sigma_{a \hat{a}}^2 \mu_{a} \mu_{\hat{a}}}{(\sigma_{a}^2+\sigma_{\hat{a}}^2)(\mu_{a}^2+\mu_{\hat{a}}^2)}
\label{eq:UIQI}
\end{equation*}
where
$\left(\mu_{a},\mu_{\hat{a}},\sigma_{a}^2,\sigma_{\hat{a}}^2\right)$
are the sample means and variances of $a$ and $\hat{a}$, and
$\sigma_{a \hat{a}}^2$ is the sample covariance of
$\left(a,\hat{a}\right)$. The range of UIQI is $[-1,1]$ and
UIQI$\left(\bfa,\hat{\bfa}\right)=1$ when $\bfa=\hat{\bfa}$. For multi-band images, the overall UIQI can be computed by averaging the UIQI computed
band-by-band.

\subsubsection{ERGAS}
The relative dimensionless global error in synthesis (ERGAS) calculates the amount of spectral
distortion in the image \cite{Wald2000}. This measure of fusion quality is defined as
\begin{equation*}
\textrm{ERGAS}=100 \times \frac{m}{n} \sqrt{\frac{1}{\nbbandima}\sum_{i=1}^{\nbbandima}\left(\frac{\textrm{RMSE}(i)}{\mu_i}\right)^2}
\end{equation*}
where $m/n$ is the ratio between the pixel sizes of the MS and HS images, $\mu_i$ is the mean of the $i$th band of the HS image, and $\nbbandima$ is the number of HS bands. The smaller ERGAS, the
smaller the spectral distortion.

\subsubsection{DD}
The degree of distortion (DD) between two images $\MATima$ and $\hat{\MATima}$ is defined as
\begin{equation*}
\textrm{DD}(\MATima,\hat{\MATima})=\frac{1}{n \nbbandima}\|\textrm{vec}(\MATima)-\textrm{vec}(\hat{\MATima})\|_1.
\end{equation*}
The smaller DD, the better the fusion.

\subsection{Comparison with other fusion methods}
\label{subsec:Pavia_comp}
This section compares the proposed fusion method with four other
state-of-the-art fusion algorithms for MS and HS images \cite{Hardie2004,Zhang2009,Yokoya2012coupled,Wei2014Bayesian}.
The parameters used for the proposed fusion algorithm have been specified as follows.
\begin{itemize}

\item The regularization parameter used in the SALSA method is $\mu = \frac{0.05}{\|{\bfN}_{\mathrm{H}}\|_F}$.
The selection of this parameter $\mu$ is still an open issue even if there are some strategies to tune it to
accelerate convergence \cite{Afonso2011}. According to the convergence theory \cite{Eckstein1992}, for any $\mu > 0$, if 
minimizing \eqref{eq:obj_lag} has a solution, say $\bfU^{\star}$, then the sequence $\left\lbrace \bfU^{(t,k)}\right\rbrace_{k=1}^{\infty}$ converges to $\bfU^{\star}$. If minimizing \eqref{eq:obj_lag} has no solution, then at least one of the sequences $\left\lbrace \bfU^{(t,k)}\right\rbrace_{k=1}^{\infty}$ or $\left\lbrace \bfG^{(t,k)}\right\rbrace_{k=1}^{\infty}$ diverges. Simulations have shown that the choice of $\mu$ does not affect significantly
the fusion performance as long as $\mu$ is positive.

\item The regularization coefficient is $\lambda=5$. The choice
of this parameter will be discussed in Section \ref{subsec:test_lambda_d}.
\end{itemize}

\begin{figure}[h!]
\centering
    \subfigure{
    \label{fig:subfig:Ref}
    \includegraphics[width=0.2\textwidth]{figures/ROSIS/Original}}
    \subfigure{
    \label{fig:subfig:HS}
    \includegraphics[width=0.2\textwidth]{figures/ROSIS/HS}}
    \subfigure{
    \label{fig:subfig:MS}
    \includegraphics[width=0.2\textwidth]{figures/ROSIS/MS}}
    \subfigure{
    \label{fig:subfig:Hardie}
    \includegraphics[width=0.2\textwidth]{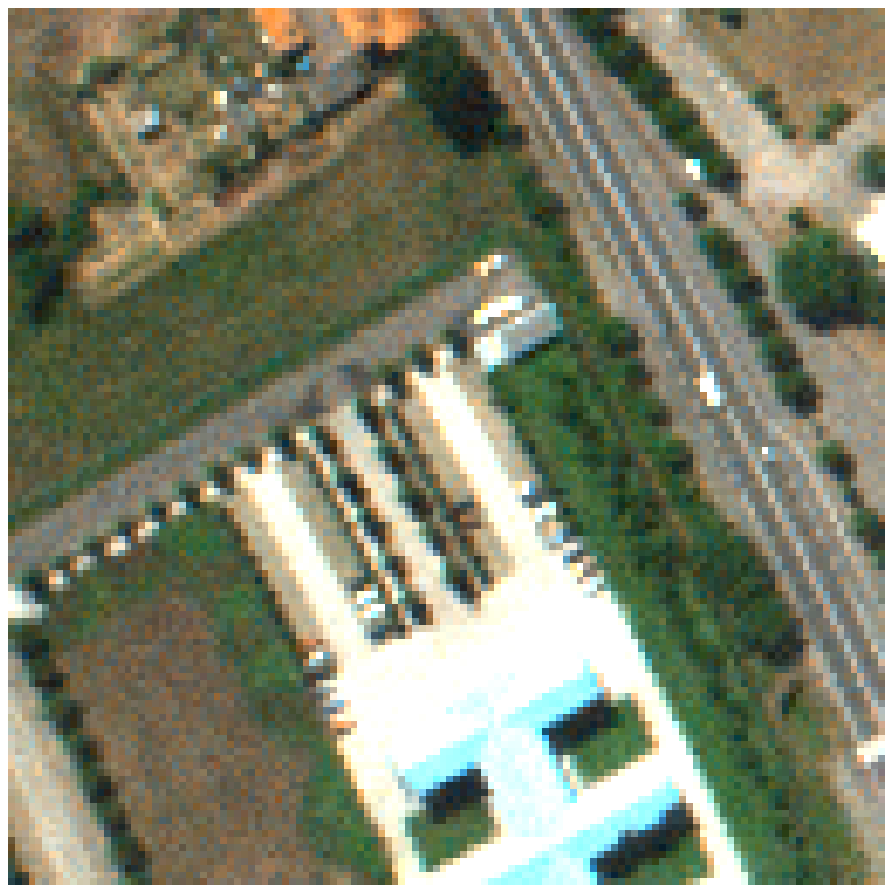}}\\
    \subfigure{
    \label{fig:subfig:Zhang}
    \includegraphics[width=0.2\textwidth]{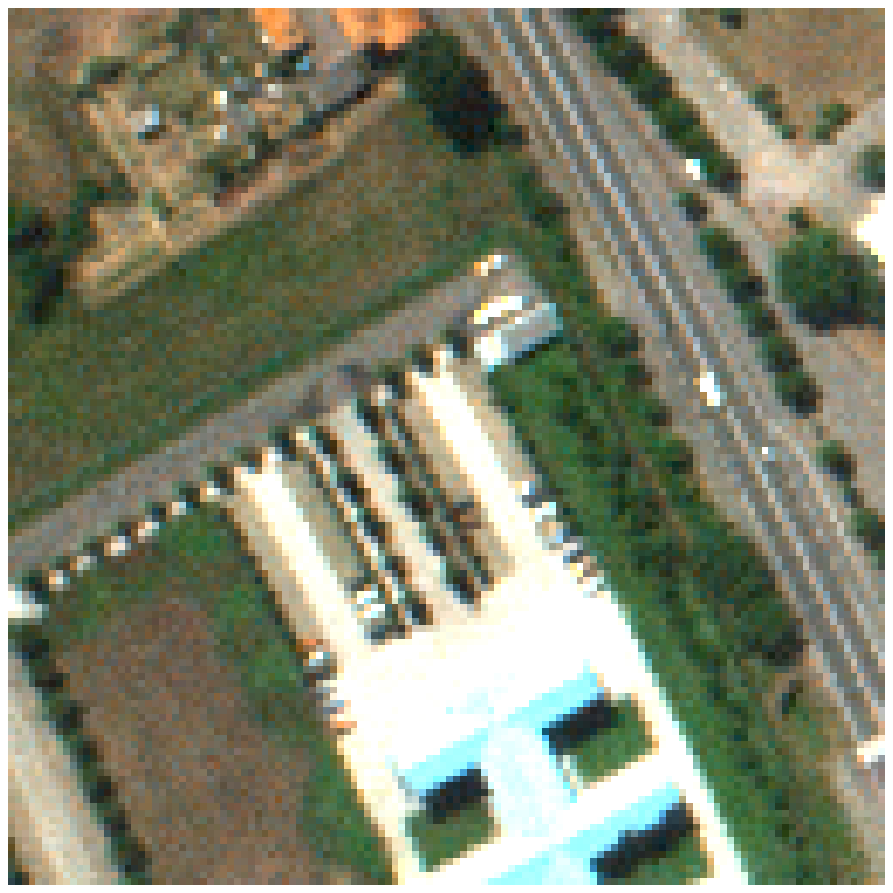}}
    \subfigure{
    \label{fig:subfig:CNMF}
    \includegraphics[width=0.2\textwidth]{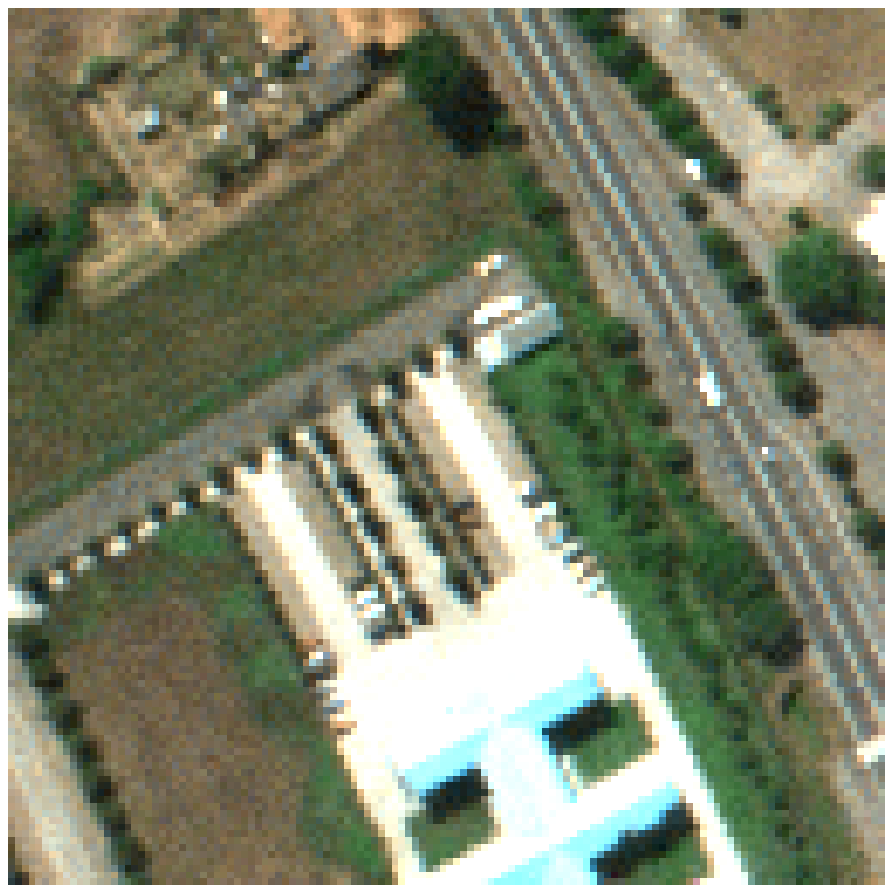}}
    \subfigure{
    \label{fig:subfig:HMC}
    \includegraphics[width=0.2\textwidth]{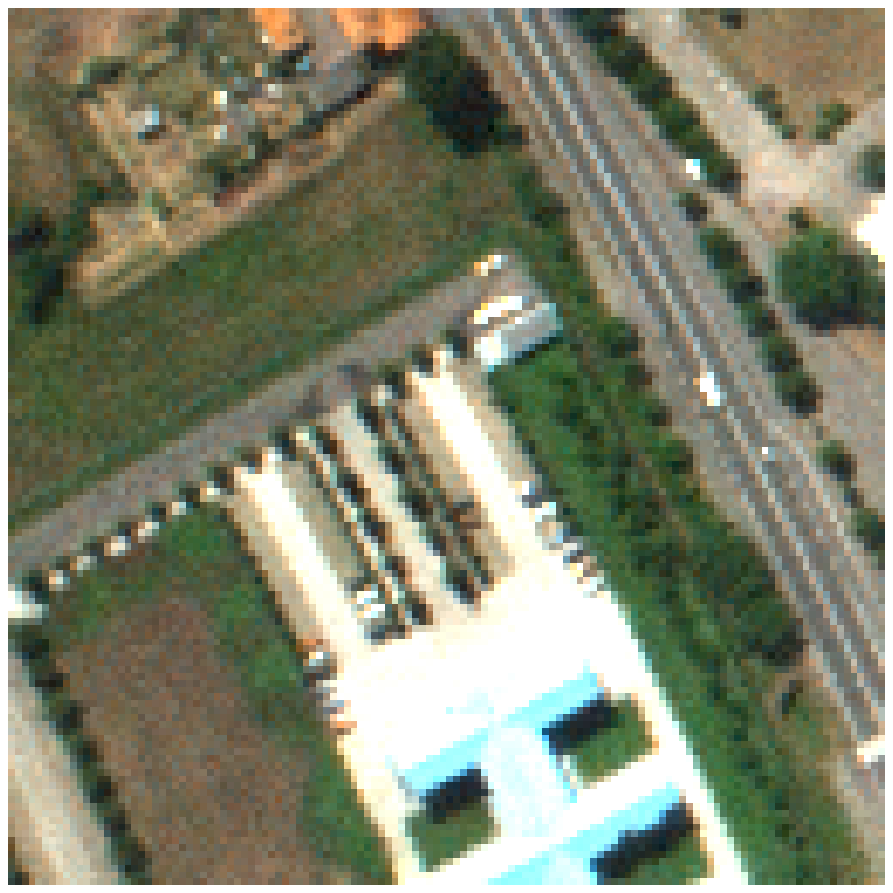}}
    \subfigure{
    \label{fig:subfig:DL}
    \includegraphics[width=0.2\textwidth]{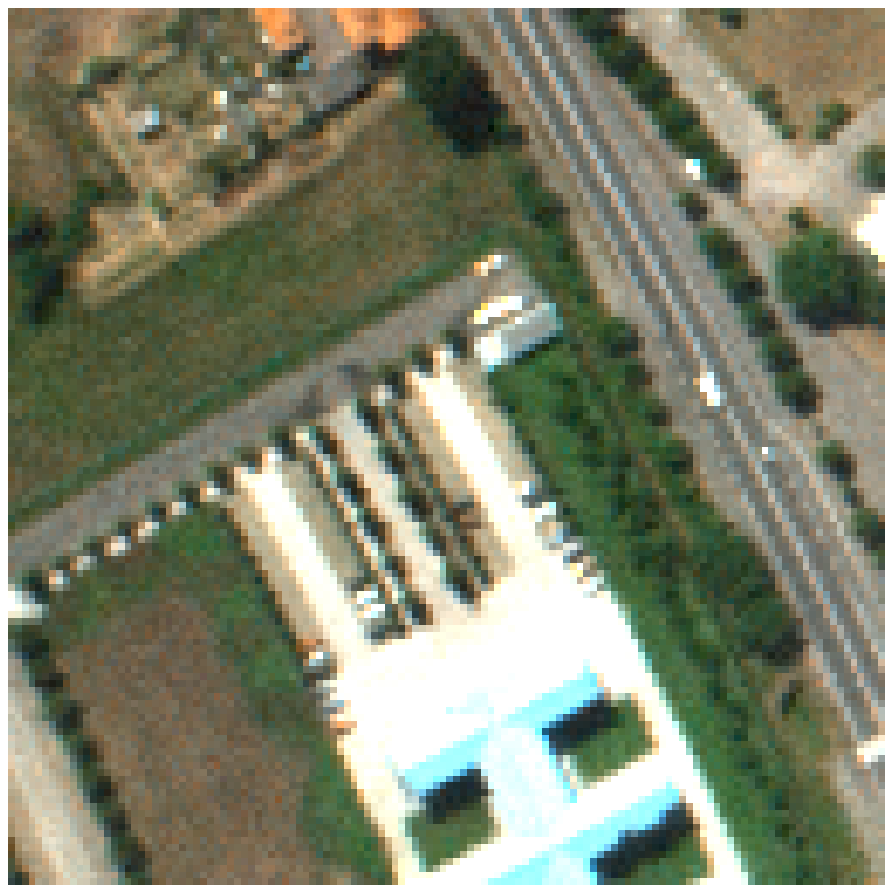}}
    \caption{Pavia dataset: (Top 1) Reference. (Top 2) HS. (Top 3) MS. (Top 4) MAP \cite{Hardie2004}. (Bottom 1) Wavelet MAP \cite{Zhang2009}. (Bottom 2) CNMF fusion \cite{Yokoya2012coupled}. (Bottom 3) MMSE estimator \cite{Wei2014Bayesian}. (Bottom 4) Proposed method.}
    \label{fig:Pavia_results}
\end{figure}

The fusion results obtained with the different algorithms are depicted 
in Fig. \ref{fig:Pavia_results}. Visually, the proposed method performs
competitively with the state-of-the-art methods. To better illustrate the 
difference of different fusion results, quantitative results are reported
in Table \ref{tb:quality_pavia} which shows the RMSE, UIQI, SAM, ERGAS and DD for
all methods. It can be seen that the proposed method always provides the best results.
It is interesting to mention that the proposed method achieves a better fusion result
with much lower computation complexity comparing with the method of HMC \cite{Wei2014Bayesian}.

\begin{table}[h!]
\centering
\caption{Performance of different MS + HS fusion methods (Pavia dataset): RMSE (in 10$^{-2}$), UIQI, SAM (in degree), 
ERGAS, DD (in 10$^{-3}$) and Time (in second).}
\begin{tabular}{c|c|c|c|c|c|c}
\hline
Methods  & RMSE   & UIQI & SAM & ERGAS & DD  & Time \\
\hline
MAP \cite{Hardie2004}   &  1.140  & 0.9876 &  1.947 &1.022 &8.608& 3\\
Wavelet MAP \cite{Zhang2009}  &  1.085  & 0.9890 &  1.777 &0.963 &8.116 & 66\\
CNMF \cite{Yokoya2012coupled}    &  1.112  & 0.9880 &  1.794 &0.999 &8.261 & $\bs{2}$\\
HMC \cite{Wei2014Bayesian}    &  0.979  & 0.9908 &  1.575 &0.877 &7.293 & 2850\\
Proposed &  $\bs{0.929}$  & $\bs{0.9916}$ & $\bs{1.470}$ & $\bs{0.831}$ & $\bs{6.843}$ & 85 \\
\hline
\end{tabular}
\label{tb:quality_pavia}
\end{table}

\subsection{Selection of the regularization parameter $\lambda$}
\label{subsec:test_lambda_d}
\begin{figure}[h!]
\centering
	\subfigure[RMSE]{
    \label{fig:subfig:RMSE}
    \includegraphics[width=0.4\textwidth]{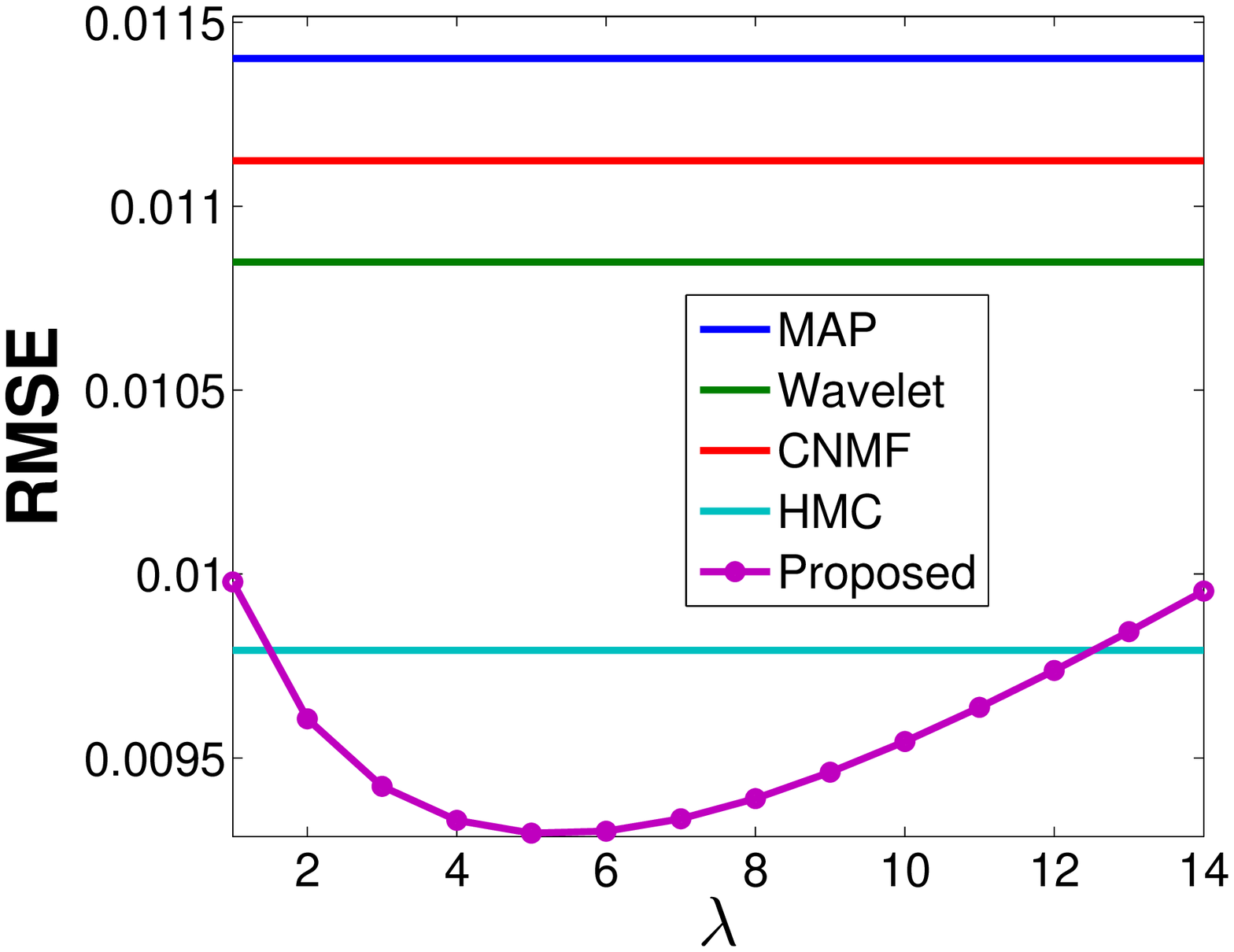}}
    \hspace{0in}
    \subfigure[UIQI]{
    \label{fig:subfig:UIQI}
    \includegraphics[width=0.4\textwidth]{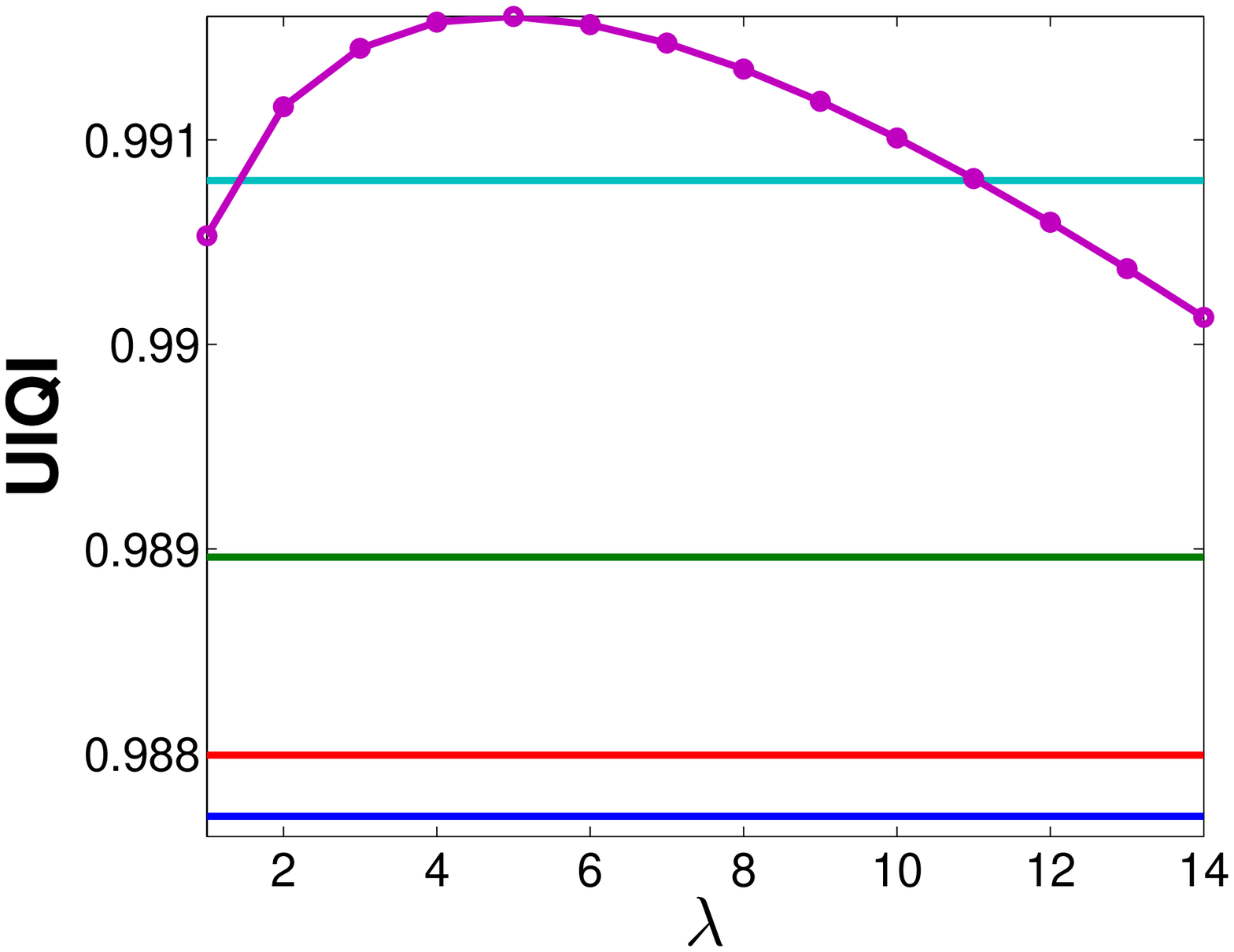}}\\
    \hspace{0in}
    \subfigure[SAM]{
    \label{fig:subfig:SAM}
    \includegraphics[width=0.4\textwidth]{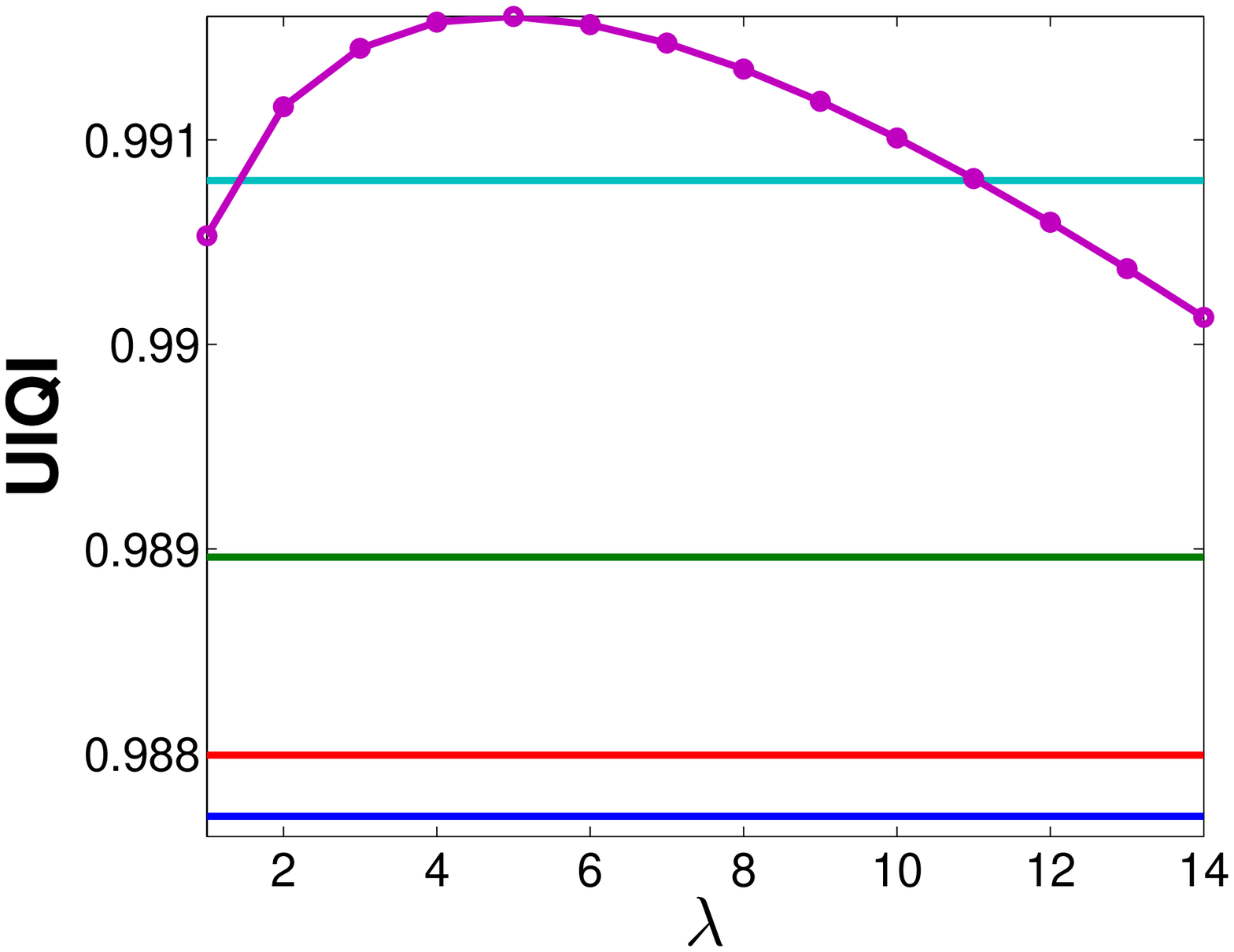}}
    \hspace{0in}
    \subfigure[DD]{
    \label{fig:subfig:DD}
    \includegraphics[width=0.4\textwidth]{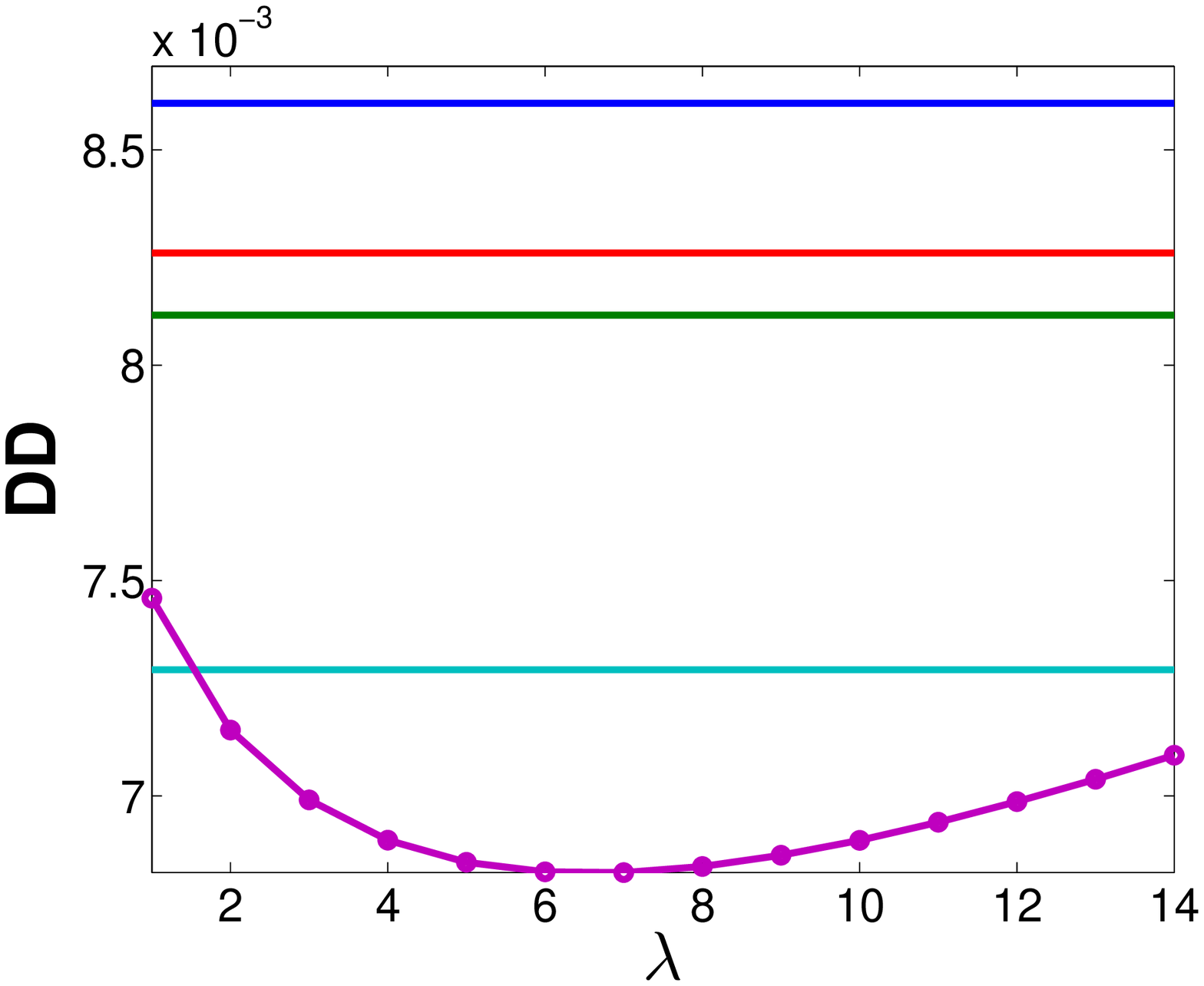}}
    \caption{Performance of the proposed fusion algorithm versus $\lambda$.}
\label{fig:lambda_d}
\end{figure}

To select an appropriate value of  $\lambda$, 
the performance of the proposed algorithm has been evaluated as a function of $\lambda$. The results
are displayed in Fig. \ref{fig:lambda_d} showing that there is no optimal value
of $\lambda$ for all the quality measures. In the simulation of Section \ref{subsec:Pavia_comp},
we have chosen $\lambda=5$ which provides the best fusion results in terms of RMSE.
It is noteworthy that in a wide range of $\lambda$, the proposed method always outperforms the 
other four methods.

\subsection{Test with other datasets}
\subsubsection{Fusion of AVIRIS data and MS data}
\label{subsec:Moffett}
\begin{figure}[h!]
\centering
\includegraphics[width=0.5\textwidth]{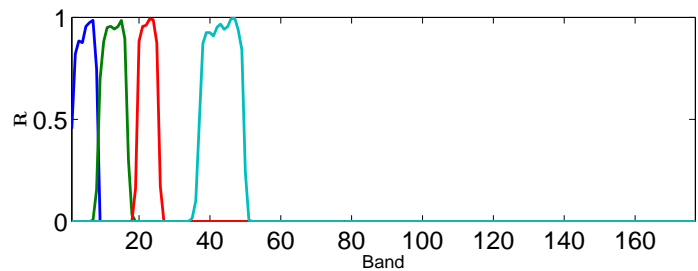}
\caption{LANDSAT spectral responses.}
\label{fig:F2_Moff}
\end{figure}

The proposed fusion method has been tested with another dataset. The reference image is a $128
 \times 128 \times 177$ hyperspectral image acquired over
Moffett field, CA, in 1994 by the JPL/NASA airborne visible/infrared
imaging spectrometer (AVIRIS) \cite{Green1998imaging}. The blurring
kernel $\bfB$, down-sampling operator $\bfS$ and SNRs for the two images are
the same as in Section \ref{subsec:prepare}. The reference image is filtered
using the LANDSAT-like spectral responses depicted in Fig. \ref{fig:F2_Moff},
to obtain a $4$-band MS image. 

\begin{figure}[h!]
\centering
    \subfigure{
    \label{fig:subfig:Ref_Moff}
    \includegraphics[width=0.2\textwidth]{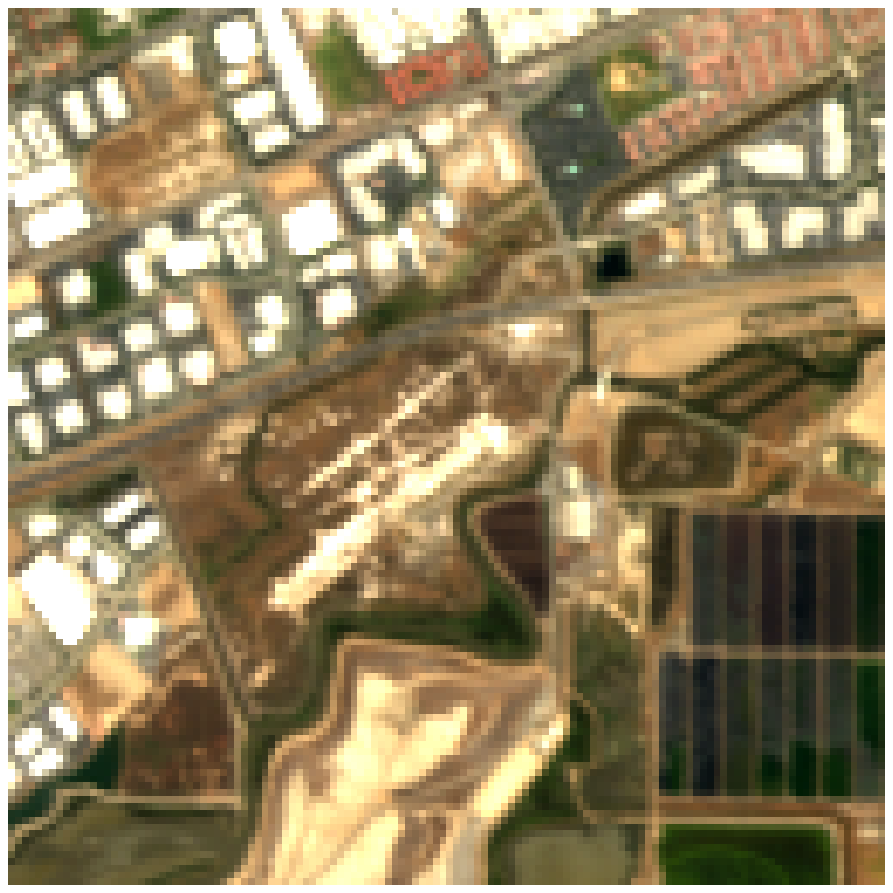}}
    \subfigure{
    \label{fig:subfig:HS_Moff}
    \includegraphics[width=0.2\textwidth]{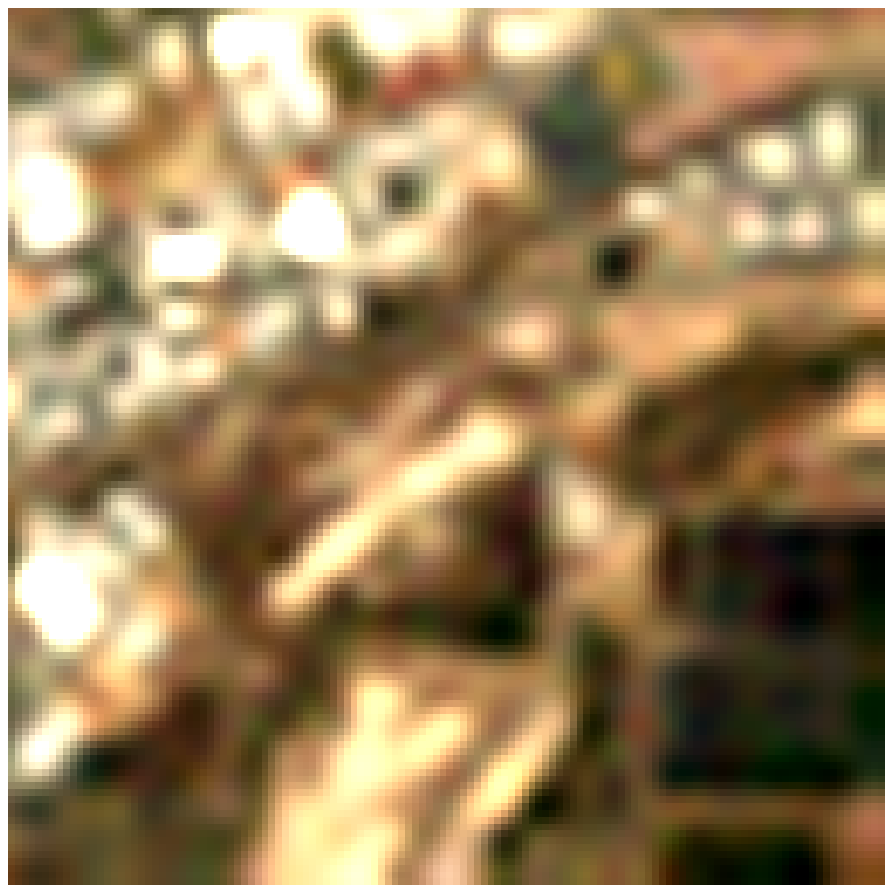}}
    \subfigure{
    \label{fig:subfig:MS_Moff}
    \includegraphics[width=0.2\textwidth]{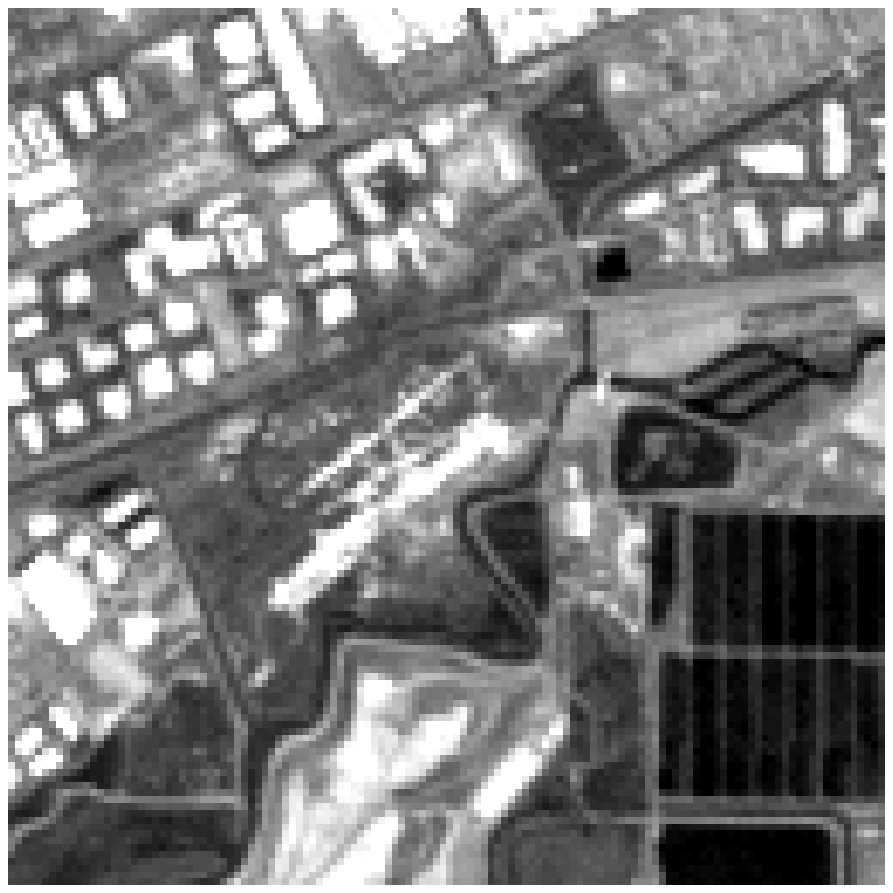}}
    \subfigure{
    \label{fig:subfig:Hardie_Moff}
    \includegraphics[width=0.2\textwidth]{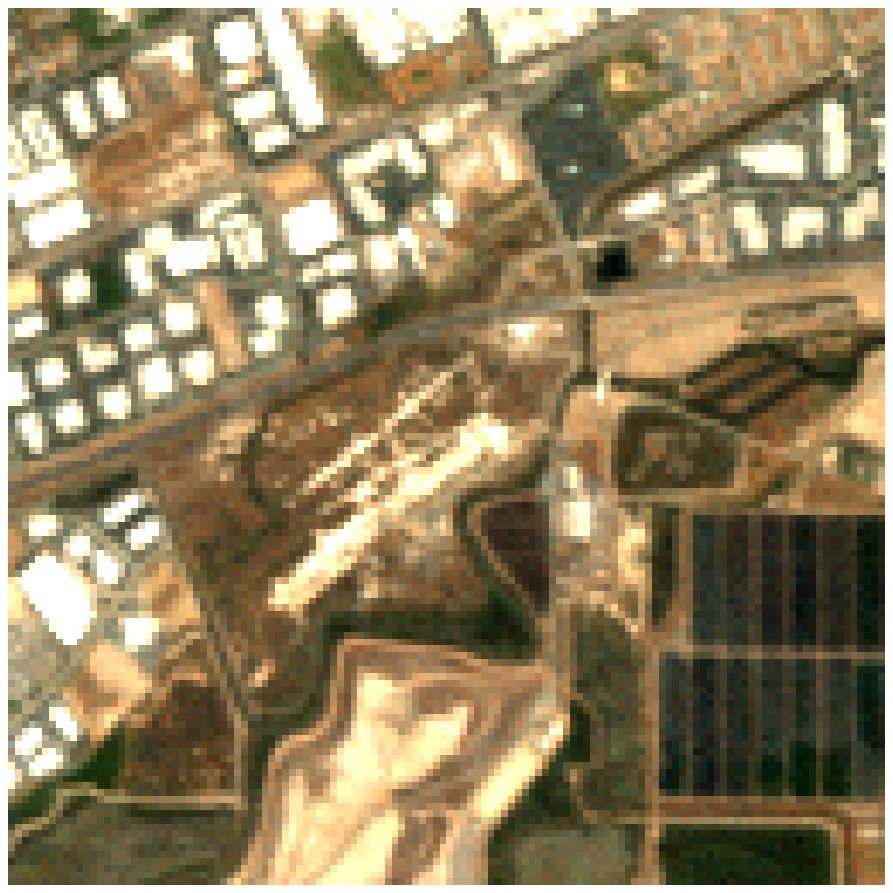}}\\
    \subfigure{
    \label{fig:subfig:Zhang_Moff}
    \includegraphics[width=0.2\textwidth]{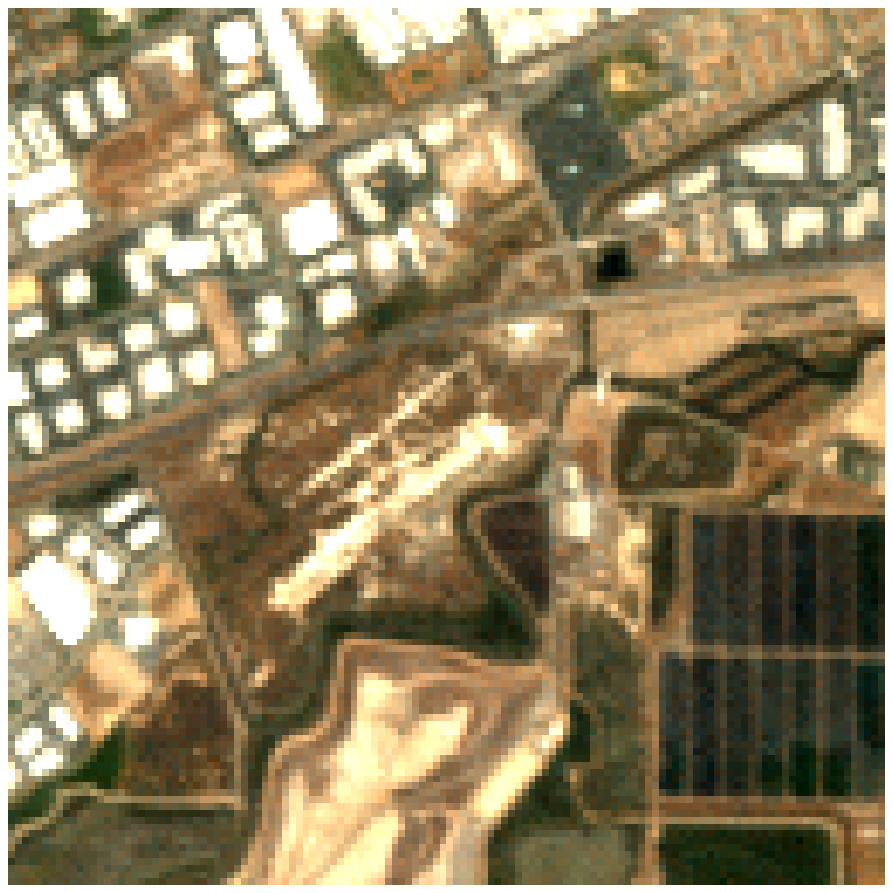}}
    \subfigure{
    \label{fig:subfig:CNMF_Moff}
    \includegraphics[width=0.2\textwidth]{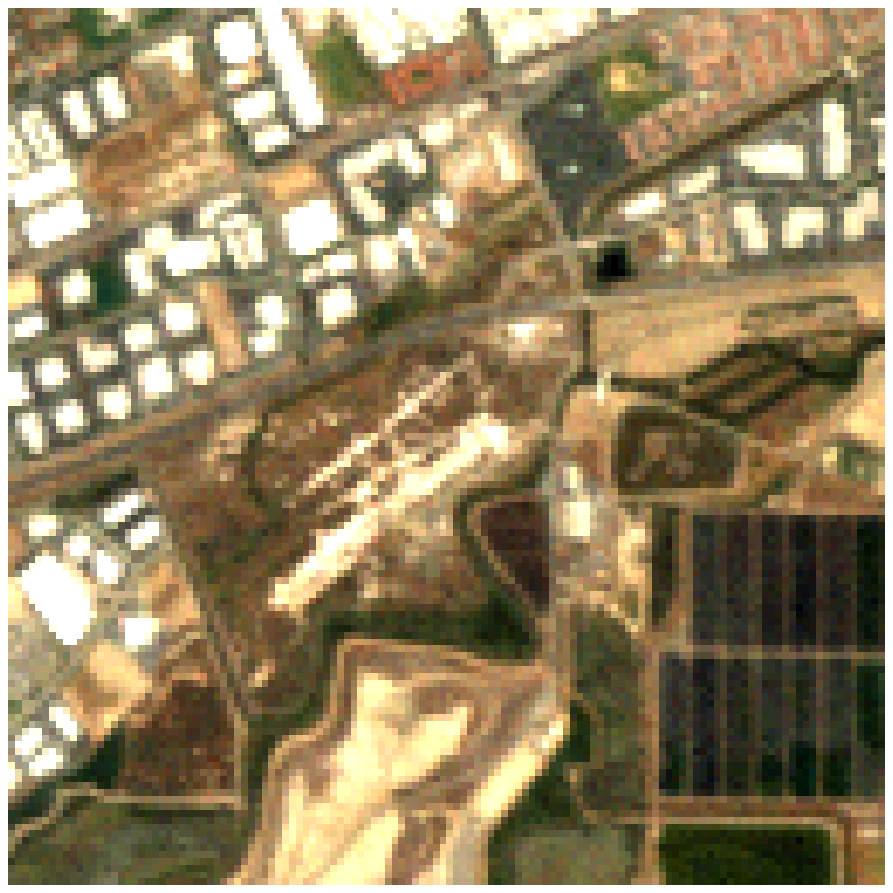}}
    \subfigure{
    \label{fig:subfig:HMC_Moff}
    \includegraphics[width=0.2\textwidth]{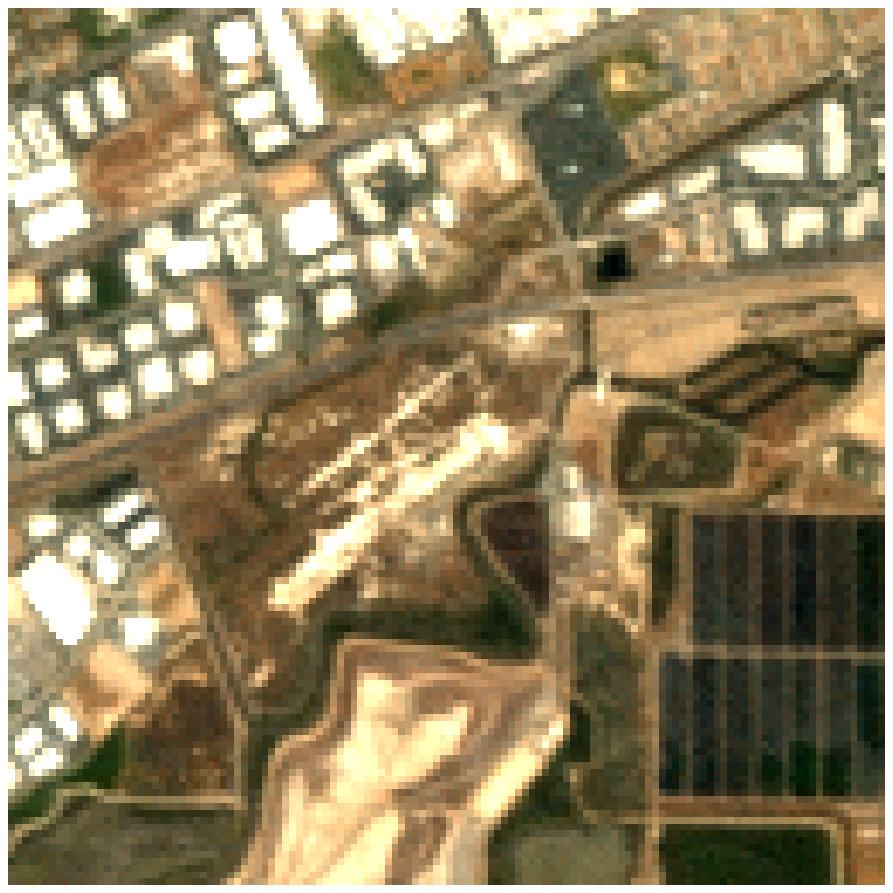}}
    \subfigure{
    \label{fig:subfig:DL_Moff}
    \includegraphics[width=0.2\textwidth]{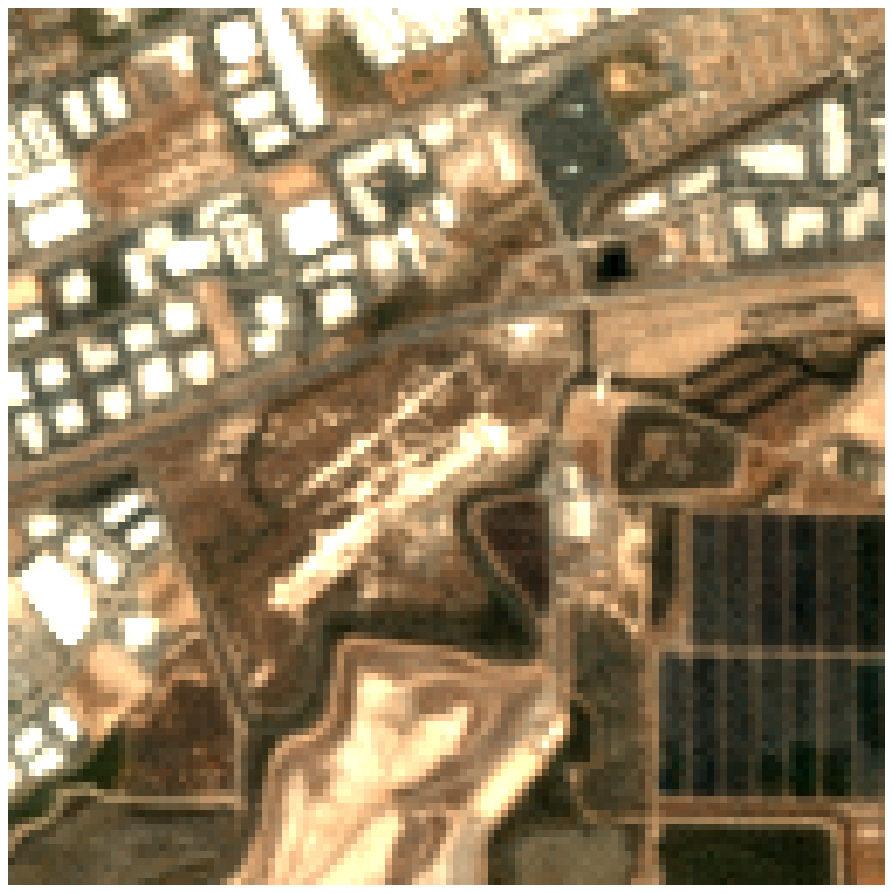}}
    \caption{Moffett dataset: (Top 1) Reference. (Top 2) HS. (Top 3) MS. (Top 4) MAP \cite{Hardie2004}. (Bottom 1) Wavelet MAP \cite{Zhang2009}. (Bottom 2) CNMF fusion \cite{Yokoya2012coupled}. (Bottom 3) MMSE estimator \cite{Wei2014Bayesian}. (Bottom 4) Proposed method.}
\label{fig:fusion_Moff}
\end{figure}

For the dictionaries and supports, the number
and size of atoms and the sparsity of the code are the same as in Section \ref{subsec:prepare}.
The proposed fusion method has been applied to the observed HS and MS
 images with a subspace of dimension $\wtm_{\lambda}=10$.
The regularization parameter has been selected by cross-validation to get the best performance
in terms of RMSE. The images (reference, MS and MS) and the fusion results obtained
with the different methods are shown in Fig. \ref{fig:fusion_Moff}. More quantitative results are reported in Table \ref{tb:quality_moff}. These results are in good agreement with what
we obtained with the previous image, proving that the proposed sparse representation 
based fusion algorithms improves the fusion quality.

\begin{table}[h!]
\centering
\caption{Performance of different MS + HS fusion methods (Moffett field): RMSE (in 10$^{-2}$), UIQI, 
SAM (in degree), ERGAS, DD (in 10$^{-2}$) and Time (in second).}
\begin{tabular}{c|c|c|c|c|c|c}
\hline
Methods  & RMSE   & UIQI & SAM & ERGAS & DD & Time \\
\hline
MAP \cite{Hardie2004}       &  2.585  & 0.9579 &  4.556 &3.111 &1.804 & $\bs{3}$\\
Wavelet MAP \cite{Zhang2009}&  2.050  & 0.9720 &  3.650 &2.466 &1.429 & 67\\
CNMF \cite{Yokoya2012coupled}    &  2.125  & 0.9696 &  3.780 &2.547 &1.477 & 12\\
HMC \cite{Wei2014Bayesian}          &  1.766  & 0.9785 &  3.205 &2.161 &1.214 & 9124\\
Proposed &  $\bs{1.668}$  & $\bs{0.9807}$ & $\bs{3.042}$ & $\bs{2.037}$ & $\bs{1.146}$ & 206
 \\
\hline
\end{tabular}
\label{tb:quality_moff}
\end{table}

\subsubsection{Pansharpening of AVIRIS data}
\label{subsec:Pansharpen}
The only difference with the Section \ref{subsec:Moffett} is that the MS image is replaced
with a PAN image obtained by averaging all the bands of the reference image (contaminated with
a Gaussian noise with SNR = $30$dB). The quantitative results are given in Table \ref{tb:quality_pan}
and are in favor of the proposed fusion method.

\begin{table}[h!]
\centering
\caption{Performance of different Pansharpening (HS + PAN) methods (Moffett field): RMSE (in 10$^{-2}$), UIQI, SAM (in degree), DD (in 10$^{-2}$) and Time (in second).}
\begin{tabular}{c|c|c|c|c|c|c}
\hline
Methods  & RMSE   & UIQI & SAM & ERGAS & DD & Time \\
\hline
MAP \cite{Hardie2004}       &  1.859  & 0.9689 &  4.162 &2.380 &1.357 & $\bs{3}$\\
Wavelet MAP \cite{Zhang2009}&  1.850  & 0.9696 &  4.186 &2.355 &1.361 & 59\\
CNMF \cite{Yokoya2012coupled}& 2.006  & 0.9650 &  4.662 &2.519 &1.478 & 4\\
HMC \cite{Wei2014Bayesian}          &  1.765  & 0.9727 &  4.074 &2.252 &1.306 & 7458\\
Proposed &  $\bs{1.734}$  & $\bs{0.9733}$ & $\bs{3.950}$ & $\bs{2.214}$ & $\bs{1.268}$ & 421 \\
\hline
\end{tabular}
\label{tb:quality_pan}
\end{table}

\section{Conclusions}
\label{sec:conclusions}
In this paper, we proposed a novel method for hyperspectral and multispectral image fusion
based on a sparse representation. The sparse representation ensured that the target image could
be well represented by the atoms of dictionaries a priori learned from the observations. 
Identifying the supports jointly with the dictionaries circumvented the difficulty inherent to sparse coding.
An alternate optimization algorithm, consisting of an alternating direction method of multipliers 
and a least square regression, was designed to minimize the target function. Compared with the other 
four state-of-the-art fusion methods, the proposed fusion method offered smaller spatial error and smaller spectral
distortion with an manageable computation complexity. This improvement was attributed to the specific sparse prior designed to regularize the resulting inverse problem. Future works include the estimation the regularization parameter $\lambda$
within the fusion scheme. Updating the dictionary jointly with the target image is also of interesting.

\section*{Acknowledgments}
The authors would like to thank Dr. Paul Scheunders and Dr. Yifan
Zhang for sharing the codes of \cite{Zhang2009}, Dr. Naoto Yokoya
for sharing the codes of \cite{Yokoya2012coupled} and Jordi Inglada,
from Centre National d'\'Etudes Spatiales (CNES), for providing the
LANDSAT spectral responses used in the experiments.

\appendix
Since $\bfU$ is a priori distributed according to a Gaussian distribution, and $\bfY_{\mathrm{M}}$ is also Gaussian distributed conditional on $\bfU$
(as the noise is i.i.d Gaussian), $\bfU$ is also Gaussian distributed conditional on $\bfY_{\mathrm{M}}$ \cite{Kay1988}.
Let define $\bfU = \left[\bsu_1,\ldots,\bsu_{n}\right]$ and $\bfY_{\mathrm{M}} = \left[\bsy_{\mathrm{M},1},\ldots,\bsy_{\mathrm{M},n}\right]$, where the vectors $\bsu_i$ represent the pixels of the target image $\bfU$ and vectors $\bsy_{\mathrm{M},i}$ represent the pixels of MS image. Assuming that the vectors $\bsu_i$ are priori spatially decorrelated,
the distribution of $\bfU$ conditional on $\bfY_{\mathrm{M}}$ is
\begin{equation*}
p(\bfU|\bfY_{\mathrm{M}})= \prod_{i=1}^{n} p(\bsu_i|\bsy_{\mathrm{M},i}) = \prod_{i=1}^{n} \calN(\bs{\mu}_{\bsu_i|\bsy_{\mathrm{M},i}},\bfC_{\bsu_i|\bsy_{\mathrm{M},i}})
\end{equation*}
We propose to define a rough approximation $\tilde{\bfU}$ of the projected target image $\bfU$
as the conditional mean $\bs{\mu}_{\bfU|\bfY_{\mathrm{M}}} =\left[ \bs{\mu}_{\bsu_1|\bsy_{\mathrm{M},1}}, \ldots, \bs{\mu}_{\bsu_n|\bsy_{\mathrm{M},n}}\right] \in \mathbb{R}^{\nbbandima \times n}$.
Straightforward computations lead to
\begin{equation}
\hat{\bs{\mu}}_{\bsu_i|\bsy_{\mathrm{M},i}}= \mathrm{E}\left[\bsu_i\right]+ \bfC_{\bsu_i,\bsy_{\mathrm{M},i}} \bfC_{\bsy_{\mathrm{M},i},\bsy_{\mathrm{M},i}}^{-1}
\left[\bsy_{\mathrm{M},i}-\mathrm{E} \left[\bsy_{\mathrm{M},i}\right]\right].
\end{equation}
It can be seen that the computation of $\hat{\bs{\mu}}_{\bsu_i|\bsy_{\mathrm{M},i}}$ requires to determine
$\mathrm{E}\left[\bsu_i\right]$, $\bfC_{\bsu_i,\bsy_{\mathrm{M},i}}$, $\bfC_{\bsy_{\mathrm{M},i},\bsy_{\mathrm{M},i}}$
and $\mathrm{E} \left[\bsy_{\mathrm{M},i}\right]$. The prior mean of projected image $\mathrm{E}\left[\bsu_i\right]$
is approximated by spatially interpolated observed HS imagery and the prior mean of MS image
$\mathrm{E} \left[\bsy_{\mathrm{M},i}\right]$ is approximated by a spatially smoothed version of
the  MS image bands. These approximates are able to
capture most of the nonstationarity exhibited by most natural images \cite{Hunt1976}. The estimation of the covariance matrix
is conducted at the lower resolution of the observed HS imagery which is quite simple. More details
about these approximations can be found in \cite{Hardie2004}.

\newpage
\bibliographystyle{ieeetran}
\bibliography{strings_all_ref,biblio_all}
\end{document}